\documentclass[11pt]{article}
\usepackage[final]{acl}
\usepackage{times}
\usepackage{latexsym}
\usepackage[T1]{fontenc}
\usepackage[utf8]{inputenc}
\usepackage{microtype}
\usepackage{inconsolata}
\usepackage{graphicx}
\usepackage{subcaption}
\usepackage{booktabs}
\usepackage{tabularx}
\usepackage{multirow}
\usepackage{amsmath,amssymb}
\usepackage{enumitem}
\usepackage{xcolor}
\usepackage{colortbl}
\definecolor{resultgreen}{RGB}{208,228,219}
\usepackage{listings}
\usepackage{placeins}
\newcommand{\method}{DSWM}

\newcommand{\withsdstack}[2]{\shortstack{#1\\{\scriptsize\textcolor{gray}{$\pm$#2}}}}

\newcommand{\withsd}[2]{\mbox{#1{\scriptsize\color{black}\ensuremath{\,\pm\,}#2}}}

\title{Anticipating Social Consequences with Distributional World Models}

\author{Viet-Thanh Pham$^1$, Thuy-Trang Vu$^1$, Zhuang Li$^2$, Lizhen Qu$^1$\thanks{Corresponding author.}, Gholamreza Haffari$^1$ \\ $^1$Department of Data Science \& AI, Monash University, Australia
\\ $^2$Royal Melbourne Institute of Technology, Australia
\\ {\small \texttt{thanh.pham1@monash.edu, trang.vu1@monash.edu, zhuang.li@rmit.edu.au,}}
\\ {\small \texttt{ lizhen.qu@monash.edu, gholamreza.haffari@monash.edu}}}

\begin{document}
\maketitle
\begin{abstract}
Social world models predict how interactions develop after an action, helping agents negotiate, cooperate and mediate. In these situations, one action can lead to several plausible futures, hence predicting only one future can overlook consequential alternatives, while exploring them through repeated simulations increases inference cost. We propose \method{}, a Distributional Social World Model that predicts alternative responses, their probabilities and immediate and final outcome values in one output per candidate action. We train a 4B model on repeated simulations using supervised fine-tuning and GRPO. At inference time, the actor uses these possible consequences to choose an action. Across three seen and two unseen environments, with interaction backbones absent from training collection, \method{} achieves the best average performance on 13 of 15 metrics. Ablations show that retaining multiple futures and their predicted probabilities improves outcomes in all five environments.

\end{abstract}
\section{Introduction}\label{sec:intro}

\begin{figure}[!t]
\centering
\includegraphics[width=\columnwidth]{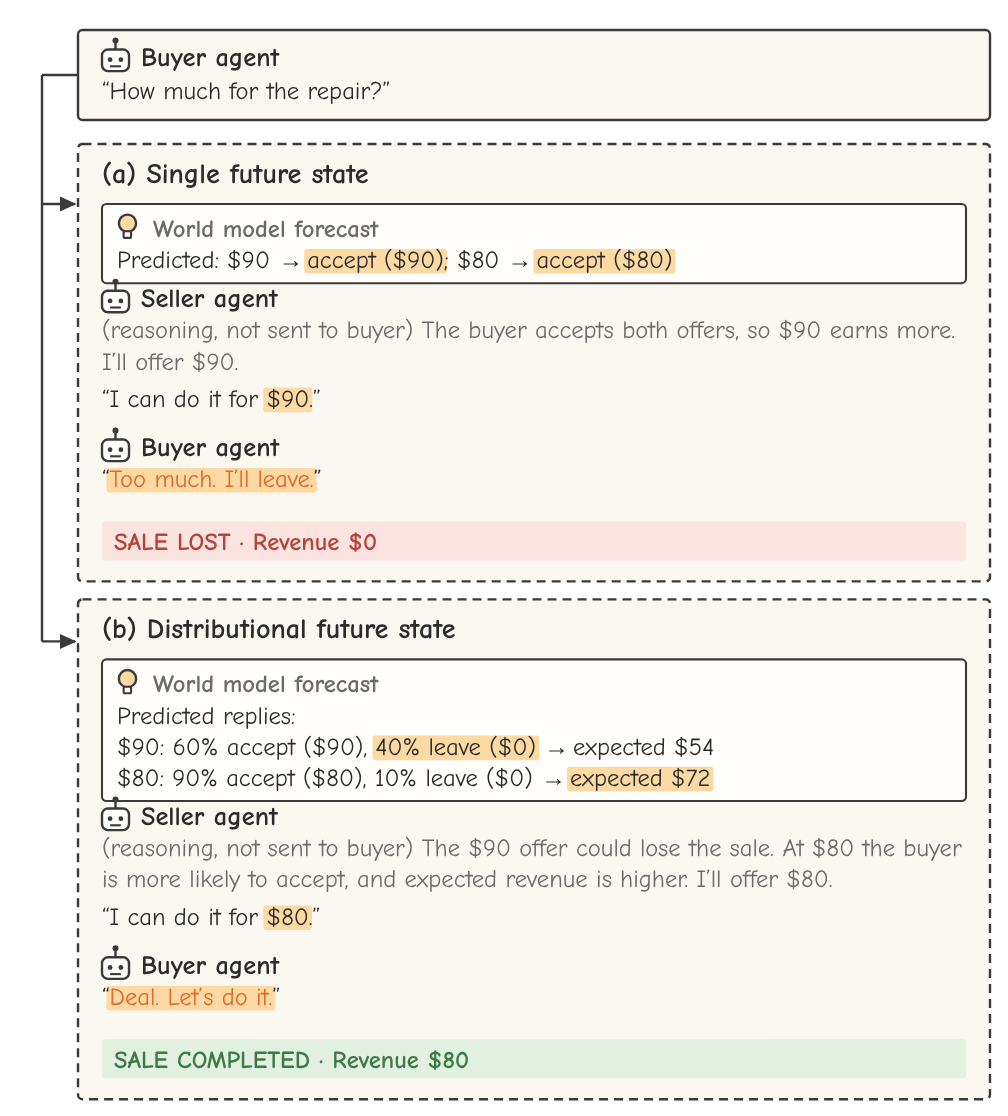}
\caption{How distributional future states guide decisions.}
\label{fig:intro}
\vspace{-2em}
\end{figure}

A \emph{social world model} describes how a social interaction may change when an agent takes an action. For example, a customer may accept an offer, counter it or leave, with different consequences for the agent. The beliefs and goals of the other parties help determine which response occurs. \citet{socialwm} describe social world models through the evolving environment, agent observations and actions, while mental world modeling couples physical changes with changes in beliefs and other mental states \citep{mentis}. By anticipating these changes before an action is taken, social world models can support negotiation, cooperation and mediation \citep{sotopia,agentsense,promediate}. Herein, outcome measures such as revenue, goal completion and cost make the consequences of different actions comparable \citep{filzmoser2019offer,kodis}.

Unlike certain physical environments that can be modeled with deterministic transitions to a single next state \citep{vjepa2}, social environments can produce several plausible future states from the same visible situation and action. Participants may have different private goals and choose different responses. A distribution over future states is therefore a natural representation: an agent should consider both what may happen and how likely it is. Figure~\ref{fig:intro} illustrates how this changes a decision. Predicting acceptance for either offer leads the seller to choose \$90, but the buyer leaves. When the prediction also represents rejection and its probability, the seller chooses \$80 for its higher acceptance probability and expected revenue, and completes the sale. However, these alternatives are not always explicit in prior approaches. S3AP Foresee-and-Act uses one predicted next state at each simulation step \citep{socialwm}, leaving alternative responses to that action implicit. GDP-Zero considers multiple future states through repeated simulations during search \citep{gdpzero}, which requires additional inference-time computation. These approaches motivate a compact output that presents alternative responses, their probabilities and their consequences together to guide decisions.

To address this limitation, we propose \method{}, a \emph{Distributional Social World Model}. Given the visible interaction and a candidate action from an agent, DSWM predicts multiple future states, 
each represented as a response cluster with a probability, a representative reply, and immediate and final outcome values.
We construct the training data from repeated simulations, resulting in a 4B world model trained with supervised fine-tuning and GRPO. At inference time, DSWM provides these alternatives in one prediction per candidate action, allowing the actor to consider different responses and their likelihoods before choosing an action. To demonstrate the effects of applying DSWM, we perform extensive evaluation on both seen and unseen environments, showing that DSWM outperforms other world modeling methods in 13 of 15 metrics. In summary, our contributions are as follows:
\begin{itemize}[leftmargin=*,itemsep=1pt,topsep=3pt]
\item An explicit distributional representation that connects alternative social responses with their likelihoods and consequences for action evaluation.
\item A training procedure that constructs targets from repeated interactions and learns them through supervised fine-tuning and GRPO.
\item An evaluation across three seen and two unseen environments with unseen agent backbones. \method{} achieves the best mean on 13 of 15 metrics; ablations examine the importance of probability assignments, response information and diverse outcome supervision.
\end{itemize}

\section{Methodology}\label{sec:method}

\begin{figure*}[t]
\centering
\includegraphics[width=\textwidth]{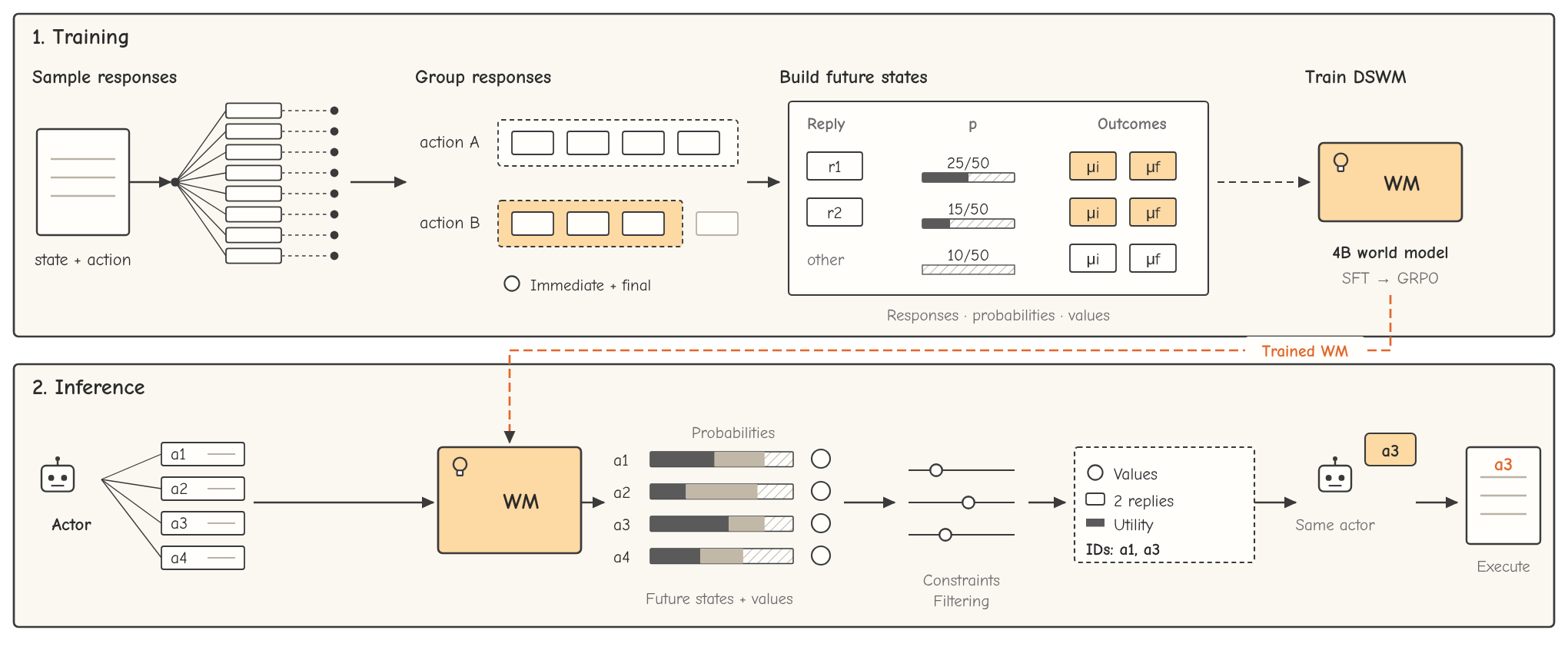}
\caption{Training and inference with \method{}: repeated simulations yield response clusters, probabilities and outcome values for training; at inference, predicted future states guide filtering and actor selection. Some of the 50 trials are drawn. Dotted branches indicate ordinary continuation for final labels; $\mu_i$ and $\mu_f$ denote immediate and final means. Values and action IDs are illustrative.}
\label{fig:method}
\vspace{-1em}
\end{figure*}

We first define the social world model and explain how its future-state representation supports action evaluation (\S\ref{sec:overview}). We then discuss how we train \method{} (\S\ref{sec:train}) and apply it at inference time (\S\ref{sec:select}), as illustrated in Figure~\ref{fig:method}.

\subsection{Social World Models}\label{sec:overview}

A social world model predicts how an interaction develops after an action, taking the responses of other agents into account. \method{} is an action-conditioned model of this kind: for one candidate action, it summarizes the possible responses of the other parties and their consequences.

\paragraph{Outcome measures.}
We use \emph{outcome measures} to describe distinct consequences, such as revenue, goal completion and privacy violations, following negotiation and dispute-resolution research \citep{filzmoser2019offer,kodis}. Their observed or predicted numbers are \emph{outcome values}. In reinforcement learning, a reward defines the optimization objective, while a $Q$-value is the expected cumulative reward after an action under a continuation policy \citep{sutton2018}. An outcome measure can supply a reward, but we keep measures separate so that the actor can weigh objectives and enforce constraints at inference. For example, revenue and privacy costs remain visible before they are combined into utility. Final expected outcome values coincide with $Q$-values when those outcomes define the return; individual cluster values additionally condition on a response.

\paragraph{Input.}
We call the agent whose actions are evaluated the \emph{focal agent}. Its \emph{actor} proposes and selects actions, while the world model (WM) evaluates their possible consequences. At a decision point, the WM receives visible context $x$, one candidate action $a$ and a query $q$. The context contains the setting, the focal agent profile and objectives, and the interaction history visible to that agent. Private goals of other parties appear only if disclosed. The query specifies the requested outcome measures and horizons (Appendix~\ref{app:spec}).

\paragraph{Output future states.}
For each candidate action, \method{} outputs a distribution over up to $K$ response clusters. A \emph{response segment} contains the messages and actions of the other parties until the focal agent next acts or the interaction ends. Each cluster represents a set of similar response segments; Section~\ref{sec:train} explains how we construct them. The output describes cluster $k$ by a short response text $r_k$, its probability $p_k$ and, for each requested measure $j$ and horizon $h$, the mean outcome value $\mu_{k,j,h}$ given a response in that cluster. The \emph{immediate} horizon evaluates the interaction after the response segment. The \emph{final} horizon evaluates it at termination under ordinary continuation, in which the actor and the other parties keep interacting under the unguided policy stated in the query. Thus, the responses and their outcome values summarize possible future states, with probabilities assigned to the response clusters. An optional residual bucket carries the probability and conditional means of the remaining responses and has no response text.

\paragraph{Relation to prior definitions.}
Like earlier social and mental world models \citep{socialwm,mentis}, \method{} predicts changes following an action. It focuses on a distributional view of response alternatives and their consequences for decisions, rather than reconstructing the complete hidden environment or private mental states. Its output is a distribution over response clusters with conditional mean outcome values, not a full joint distribution of those values.

\paragraph{Decision-theoretic grounding.}
To compare actions, we average the outcome values across the response clusters of each action:
\begin{equation}\label{eq:expected}
\hat{v}_{j,h}(a)=\sum_k p_k\,\mu_{k,j,h},
\end{equation}
where the sum includes the residual bucket when present. This is the law of total expectation: the overall expected value is the probability-weighted average of the conditional values. In the lower panel of Figure~\ref{fig:intro}, the \$90 offer has expected revenue $0.6\times90+0.4\times0=54$, while the \$80 offer yields $0.9\times80+0.1\times0=72$. The seller therefore chooses \$80: accounting for rejection reverses the preference based on acceptance alone.

The representation therefore connects to expected-utility theory, which evaluates an action by weighting the utility of possible outcomes by their probabilities \citep{vonneumann1944}. For one measure or a fixed weighted sum, utility is linear in the outcome values, so accurate expected values suffice to compare actions. When an objective combines measures nonlinearly, utility computed from cluster means only approximates expected utility, because the representation omits within-cluster variation and covariance (\S\ref{sec:select}). Predicted values are also approximate. The response descriptions show which responses produce each average, and our analyses test whether this information improves decisions beyond the numerical values. To estimate these training targets, we repeatedly take each action from the same saved state.

\subsection{Training DSWM}\label{sec:train}\label{sec:collect}\label{sec:modes}

\paragraph{Collecting future states.}
We save the complete environment state at a focal decision point, execute the same candidate action from that state $N{=}50$ times, and sample the responses of the other parties. In batches selected for final-outcome collection, every sampled interaction continues to termination under the ordinary continuation policy; other batches supply immediate labels only (Appendix~\ref{app:method-collection}). Repeating an action and averaging the resulting outcomes follows Monte Carlo action evaluation \citep[Ch.~5]{sutton2018}. Collection can use the saved full state, while the WM learns to predict from the visible context alone.

We then measure the consequences of each sampled response and, when available, its completed interaction. Deterministic environments compute measures such as negotiation utility and revenue directly. Otherwise, an LLM judge applies benchmark rubrics or additional measures such as partner trust and secret leakage (Appendix~\ref{app:data}).

\paragraph{Constructing distributional targets.}
To summarize the samples, we first partition responses by their representative actions, such as acceptance or counteroffers, so that these remain distinct even when their wording is similar. We then cluster semantically similar texts within each partition. The $K$ largest clusters each retain a representative sampled response, and the remaining clusters form the residual bucket (Appendix~\ref{app:method-modes}). For cluster $C_k$, with observed outcome values $s_{i,j,h}$ for member sample $i$, the targets are
\begin{equation}\label{eq:targets}
p_k=\frac{|C_k|}{N},\qquad
\mu_{k,j,h}=\frac{1}{|C_k|}\sum_{i\in C_k}s_{i,j,h}.
\end{equation}
The probabilities are empirical frequencies under the policies. All measures share one response partition and are averaged separately within each cluster, so binary measures can have fractional conditional means. We request a measure and horizon only when labels exist for all $N$ samples, which for final values means batches continued to termination. Consequently, Eq.~\ref{eq:expected} recovers the sample mean before rounding: clustering changes the representation while preserving expected outcome values.

\paragraph{Supervised fine-tuning.}
We serialize these targets as JSON containing response texts, probabilities and conditional mean outcome values, and train the WM to generate them with SFT. Inputs contain the visible context, candidate action and query, excluding reference continuations, observed labels and hidden answer keys. Query views vary the requested measures, horizons and cluster count (Appendix~\ref{app:data}). This teaches the WM to express alternative future states in a common output format.

\paragraph{Reinforcement learning.}
SFT can penalize a prediction that describes the same samples with different wording or grouping. We therefore continue training with GRPO \citep{grpo}, evaluating generated distributions against the raw reference continuations. Each reference response is matched to its most similar predicted response when their representative actions agree and semantic similarity reaches a threshold; otherwise, it counts as missed. The reward is
\begin{equation}\label{eq:reward}
\begin{split}
R=g\,(&0.35R_{\mathrm{dist}}+0.20R_{\mathrm{cov}}+0.10R_{\mathrm{prec}}\\
      &+0.30R_{\mathrm{score}}+0.05R_{\mathrm{div}}).
\end{split}
\end{equation}
The gate $g$ is zero for an output that cannot be parsed, omits a requested measure or has the wrong cluster count. $R_{\mathrm{dist}}$ is one minus the total-variation distance between predicted probabilities and matched reference frequencies. $R_{\mathrm{cov}}$ measures reference mass matched to explicit responses, while $R_{\mathrm{prec}}$ measures predicted mass receiving at least one reference. $R_{\mathrm{score}}$ is one minus the mean squared normalized error between each reference outcome and the conditional mean of its assigned cluster. Finally, $R_{\mathrm{div}}$ penalizes near-duplicate responses. SFT and GRPO draw on the same training pool (Appendix~\ref{app:method-training}). Together, these objectives train the probabilities, response coverage and outcome values used at inference time.

\subsection{Applying DSWM at Inference Time}\label{sec:select}

At each decision point, the actor proposes up to four actions, and invalid or duplicate proposals are removed. The trained WM then predicts response clusters for each candidate once, requesting four clusters, or one for an action that ends the interaction. From these predictions, we compute final expected outcome values using Eq.~\ref{eq:expected} and a combined utility for each candidate. For SOTOPIA-ToM, whose objective combines measures nonlinearly, we compute utility from the conditional means within each cluster and then average it by probability; the other environments use linear combinations.

We then filter candidates in two steps. First, we remove candidates whose predicted constraint costs exceed the lowest predicted cost by more than a small margin. Then, we remove a candidate if another is at least as good on every active objective and constraint and better on at least one (Figure~\ref{fig:selection}, appendix).

The actor receives the candidate actions with numerical and response guidance. Numerical guidance consists of immediate and final expected outcome values, plus combined utilities for eligible candidates. Response guidance consists of the two most probable response descriptions with their probabilities and the residual probability mass. Given this guidance and the eligible action IDs, the actor chooses an eligible action, which is executed unchanged. If all predictions are invalid, the actor selects without model guidance. Selection repeats at later turns, while the predicted final values assume ordinary continuation (Appendix~\ref{app:selection}). Thus, the WM supplies possible future states and their consequences, while the actor remains responsible for the decision.

\section{Experiments}\label{sec:experiments}

In the below sections, we examine how distributional future states support social decisions. We evaluate action selection on held-out scenario families in three training environments and two unseen environments. We then examine the contributions of alternative responses, their probabilities and their consequences, and whether learning these together reduces the cost of evaluating multiple futures.

\subsection{Benchmarks}\label{sec:benchmarks}

Five benchmark adaptations cover negotiation, social interaction, mediation and information sharing:

\paragraph{Magentic-NL}
is a two-party negotiation adapted from the restaurant and home-contractor scenarios in Magentic Marketplace \citep{magentic}. Customers seek useful purchases at favorable prices, while businesses seek revenue.

\paragraph{SOTOPIA}
tests open-ended social role-play between two characters with individual goals \citep{sotopia}. Evaluation covers goal achievement and social consequences, including relationships and adherence to social rules.

\paragraph{AgentSense}
tests goal completion and information reasoning in scenes derived from film and television scripts, spanning personal, workplace and public interactions \citep{agentsense}.

\paragraph{ProMediate}
tests multi-party mediation across healthcare, environmental policy and business cases \citep{promediate}. Here, the focal agent helps parties with competing preferences reach consensus.

\paragraph{SOTOPIA-ToM}
tests coordination and privacy-aware information sharing among three to five agents with different private knowledge and public or private communication channels \citep{sotopiatom}. Together with ProMediate, it is unseen to the trained WM.

\subsection{Training Data}\label{sec:data}

From the three training environments, we obtain 12,247 training query rows: 6,041 from Magentic-NL, 4,810 from SOTOPIA and 1,396 from AgentSense. Different query rows can reuse the same context, candidate action and sampled continuations. We follow official splits where available and otherwise hash scenario families into approximately 70/10/20 train/validation/test partitions. We also apply filtering to remove non-finite or out-of-range labels. The validation and test files contain 2,137 and 2,731 rows. Appendix~\ref{app:data} provides the exact pool composition and held-out outcome-measure splits.

\subsection{Experimental Setup}\label{sec:setup}

\paragraph{Models and evaluation.}
For these experiments, the WM is Qwen3-4B-Instruct-2507~\citep{qwen3technicalreport}. Rollout agents used to create training data use Ornith-1.5-9B~\cite{ornith_1_5}, Qwen3.5-9B~\citep{qwen3.5} and Gemma~4~12B~\citep{gemmateam2026gemma4}, whereas evaluation pools the disjoint set of Ministral~3~14B, MiniMax~H3 and Claude~Sonnet~5.
Evaluation outcome values come from the environment for deterministic metrics or from GPT-5.6-Sol for metrics evaluated by an LLM judge (Appendix~\ref{app:benchmarks}).

\begin{table*}[!t]
\centering
\footnotesize
\setlength{\tabcolsep}{3.5pt}
\begin{tabular}{@{}l ccc cc cc@{}}
\toprule
 & \multicolumn{3}{c}{Magentic-NL} & \multicolumn{2}{c}{SOTOPIA} & \multicolumn{2}{c}{AgentSense} \\
\cmidrule(lr){2-4}\cmidrule(lr){5-6}\cmidrule(lr){7-8}
Method & Cust.\ util. & Serv.\ rev. & Needs met & Goal & Overall & Goal & Info.\ acc. \\
\midrule
Actor alone        & \withsd{179.98}{8.30}  & \withsd{116.84}{7.34} & \withsd{0.35}{0.021} & \withsd{6.47}{0.24} & \withsd{2.45}{0.15} & \withsd{0.66}{0.027} & \withsd{0.86}{0.025} \\
Candidates only    & \withsd{195.46}{11.44} & \withsd{125.20}{8.51} & \withsd{0.37}{0.023} & \withsd{6.77}{0.19} & \withsd{2.48}{0.11} & \withsd{0.66}{0.024} & \withsd{0.88}{0.027} \\
\midrule
Single future      & \withsd{206.05}{8.20}  & \withsd{125.15}{8.29} & \withsd{0.37}{0.028} & \withsd{6.49}{0.23} & \withsd{2.63}{0.13} & \withsd{0.67}{0.019} & \withsd{\underline{0.91}}{0.026} \\
Direct expectation & \withsd{208.68}{12.20} & \withsd{129.33}{6.05} & \withsd{0.36}{0.030} & \withsd{7.59}{0.23} & \withsd{2.49}{0.14} & \withsd{0.70}{0.022} & \withsd{0.88}{0.029} \\
Sampled futures    & \withsd{255.56}{9.65}  & \withsd{150.01}{6.65} & \withsd{0.38}{0.024} & \withsd{7.10}{0.23} & \withsd{2.92}{0.15} & \withsd{0.59}{0.021} & \withsd{0.80}{0.024} \\
\midrule
S3AP F\&A          & \withsd{188.03}{8.73}  & \withsd{118.33}{6.24} & \withsd{0.37}{0.032} & \withsd{7.31}{0.19} & \withsd{2.83}{0.13} & \withsd{\underline{0.71}}{0.025} & \cellcolor{resultgreen}\withsd{\textbf{0.95}}{0.023} \\
GDP-Zero           & \withsd{239.25}{11.27} & \withsd{\underline{161.97}}{8.30} & \withsd{0.45}{0.030} & \withsd{\underline{8.00}}{0.18} & \withsd{2.35}{0.14} & -- & -- \\
\midrule
\method{} (SFT)         & \withsd{\underline{261.28}}{11.54} & \withsd{152.43}{8.85} & \withsd{\underline{0.49}}{0.026} & \withsd{7.83}{0.21} & \withsd{\underline{2.98}}{0.11} & \withsd{0.67}{0.025} & \withsd{0.88}{0.029} \\
\method{} (SFT + GRPO)  & \cellcolor{resultgreen}\withsd{\textbf{272.82}}{7.91} & \cellcolor{resultgreen}\withsd{\textbf{170.64}}{7.48} & \cellcolor{resultgreen}\withsd{\textbf{0.51}}{0.025} & \cellcolor{resultgreen}\withsd{\textbf{8.21}}{0.20} & \cellcolor{resultgreen}\withsd{\textbf{3.24}}{0.15} & \cellcolor{resultgreen}\withsd{\textbf{0.72}}{0.023} & \withsd{0.88}{0.024} \\
\bottomrule
\end{tabular}
\caption{Held-out scenario families of the training environments, evaluated with backbones absent from training rollouts. Mean $\pm$ SD over five seeds, each averaging three backbones; higher is better on every metric. Customer utility and service revenue are reported for the corresponding focal role. Green shading and bold mark the best mean; underlining marks the second-best distinct mean. Ties share formatting. GDP-Zero applies only to two-party settings.}
\label{tab:main-seen}
\end{table*}
\begin{figure*}[!t]
\centering
\includegraphics[width=\textwidth]{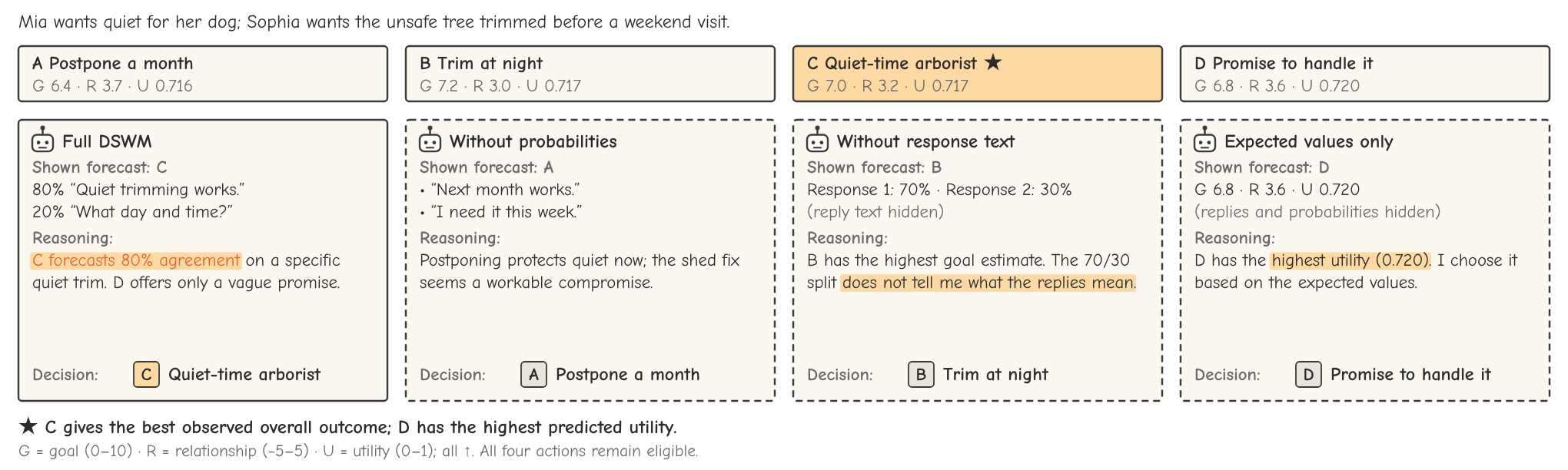}
\caption{SOTOPIA choices and recorded reasoning with expected values and eligibility fixed. C achieves the best observed overall outcome. Two clusters (main evaluation: four); text abbreviated. Details: Appendix~\ref{app:illustrated-display}.}
\label{fig:decision-example}
\end{figure*}

\paragraph{Baselines.}
Without WM predictions, \emph{Actor alone} generates an action directly, while \emph{Candidates only} selects among generated actions.

We compare three prediction methods. \emph{Single future} predicts one response and its outcome values; \emph{Direct expectation} predicts expected outcome values; \emph{Sampled futures} averages outcome values from 30 independent continuations per action. All share the actor, selection procedure, training rollouts and SFT budget. Since these baselines use SFT only, we report both SFT-only and SFT + GRPO \method{}.

We also follow released implementations of S3AP Foresee-and-Act \citep{socialwm} and GDP-Zero \citep{gdpzero}, using GDP-Zero in two-party settings. Appendix~\ref{app:baselines} details configurations and utility aggregation.

\paragraph{Ablations.}
To examine why distributional outputs help, we first retain only the most likely response, testing the importance of alternative futures. We then assign uniform probabilities or perturb their weights to test how likelihoods affect decisions. These interventions recompute expected values, utilities and eligibility. Separately, removing response text or showing expected values only tests what response information adds with numerical guidance and filters fixed. Removing filters or selecting the highest-utility action automatically tests the roles of eligibility and actor selection. These variants share a checkpoint within each seed. Finally, retraining with 5--25 outcome measures under matched supervision tests whether learning diverse consequences improves transfer (Appendices~\ref{app:mechanism-analysis} and~\ref{app:sensitivity}).

\begin{table*}[t]
\centering
\footnotesize
\setlength{\tabcolsep}{4pt}
\begin{tabularx}{\textwidth}{@{}X ccc ccccc@{}}
\toprule
 & \multicolumn{3}{c}{\textbf{ProMediate}} & \multicolumn{5}{c}{\textbf{SOTOPIA-ToM}} \\
\cmidrule(lr){2-4}\cmidrule(lr){5-9}
\textbf{Method} & Cons.\ chg. & Topic eff. & Med.\ intel. & InfoMgmt & DA & IA & CPV\,$\downarrow$ & EFF \\
\midrule
Actor alone & \withsdstack{0.0576}{0.0058} & \withsdstack{0.0030}{0.0004} & \withsdstack{3.2179}{0.200} & \withsdstack{0.1769}{0.0174} & \withsdstack{0.3538}{0.0205} & \withsdstack{0.1125}{0.0138} & \withsdstack{0.3487}{0.0218} & \withsdstack{0.6961}{0.0294} \\
Candidates only & \withsdstack{\underline{0.0616}}{0.0053} & \withsdstack{0.0031}{0.0004} & \withsdstack{3.3219}{0.153} & \withsdstack{0.1789}{0.0126} & \withsdstack{0.3764}{0.0227} & \withsdstack{0.1151}{0.0153} & \withsdstack{0.3689}{0.0179} & \withsdstack{0.7525}{0.0328} \\
\midrule
Single future & \withsdstack{0.0452}{0.0039} & \withsdstack{0.0028}{0.0003} & \withsdstack{3.2051}{0.137} & \withsdstack{0.1795}{0.0115} & \withsdstack{0.3692}{0.0244} & \withsdstack{0.1278}{0.0127} & \withsdstack{0.3475}{0.0282} & \withsdstack{0.8154}{0.0313} \\
Direct expectation & \withsdstack{0.0440}{0.0050} & \withsdstack{0.0034}{0.0003} & \withsdstack{3.2919}{0.136} & \withsdstack{0.1840}{0.0133} & \withsdstack{0.4736}{0.0274} & \withsdstack{0.1352}{0.0147} & \withsdstack{0.3376}{0.0177} & \withsdstack{0.8813}{0.0391} \\
Sampled futures & \withsdstack{0.0569}{0.0054} & \withsdstack{0.0027}{0.0003} & \withsdstack{3.2914}{0.189} & \withsdstack{0.2160}{0.0146} & \withsdstack{0.4650}{0.0192} & \withsdstack{0.1249}{0.0145} & \withsdstack{0.3071}{0.0209} & \withsdstack{0.8509}{0.0382} \\
\midrule
S3AP F\&A & \withsdstack{0.0503}{0.0058} & \withsdstack{\underline{0.0036}}{0.0004} & \withsdstack{3.4387}{0.136} & \withsdstack{0.1930}{0.0126} & \withsdstack{0.4179}{0.0188} & \withsdstack{\underline{0.1477}}{0.0140} & \withsdstack{0.3412}{0.0187} & \withsdstack{0.8847}{0.0367} \\
GDP-Zero & -- & -- & -- & -- & -- & -- & -- & -- \\
\midrule
\method{} (SFT only) & \withsdstack{0.0613}{0.0061} & \withsdstack{0.0034}{0.0004} & \withsdstack{\underline{4.0129}}{0.141} & \withsdstack{\underline{0.2173}}{0.0125} & \withsdstack{\underline{0.4932}}{0.0216} & \withsdstack{0.1474}{0.0135} & \cellcolor{resultgreen}\withsdstack{\textbf{0.2281}}{0.0176} & \withsdstack{\underline{0.8859}}{0.0256} \\
\method{} (SFT + GRPO) & \cellcolor{resultgreen}\withsdstack{\textbf{0.0628}}{0.0047} & \cellcolor{resultgreen}\withsdstack{\textbf{0.0038}}{0.0004} & \cellcolor{resultgreen}\withsdstack{\textbf{4.3321}}{0.185} & \cellcolor{resultgreen}\withsdstack{\textbf{0.2508}}{0.0119} & \cellcolor{resultgreen}\withsdstack{\textbf{0.5398}}{0.0243} & \cellcolor{resultgreen}\withsdstack{\textbf{0.1695}}{0.0128} & \withsdstack{\underline{0.2450}}{0.0173} & \cellcolor{resultgreen}\withsdstack{\textbf{0.9138}}{0.0300} \\
\bottomrule
\end{tabularx}
\caption{Transfer to unseen environments and backbones; aggregation follows Table~\ref{tab:main-seen}. Mean $\pm$ s.d.\ over seeds. Higher is better except CPV; highlighting follows Table~\ref{tab:main-seen}. GDP-Zero is inapplicable to these multi-party settings.}
\label{tab:main-unseen}
\end{table*}

\begin{table*}[!tp]
\centering
\small
\setlength{\tabcolsep}{3pt}
\begin{tabularx}{\textwidth}{@{}X ccccc@{}}
\toprule
 & Magentic-NL & SOTOPIA & AgentSense & ProMediate & SOTOPIA-ToM \\
Selection variant & Cust.\ util. $\uparrow$ & Goal $\uparrow$ & Goal $\uparrow$ & Cons.\ chg. $\uparrow$ & InfoMgmt $\uparrow$ \\
\midrule
Full \method{} & \cellcolor{resultgreen}\withsd{\textbf{272.82}}{7.91} & \cellcolor{resultgreen}\withsd{\textbf{8.21}}{0.20} & \cellcolor{resultgreen}\withsd{\textbf{0.72}}{0.023} & \cellcolor{resultgreen}\withsd{\textbf{0.0628}}{0.0047} & \cellcolor{resultgreen}\withsd{\textbf{0.2508}}{0.0119} \\
Without response text & \cellcolor{resultgreen}\withsd{\textbf{272.82}}{7.91} & \withsd{8.02}{0.25} & \withsd{0.63}{0.025} & \withsd{0.0574}{0.0044} & \withsd{0.2348}{0.0112} \\
Expected values only & \cellcolor{resultgreen}\withsd{\textbf{272.82}}{7.91} & \withsd{7.98}{0.19} & \withsd{0.60}{0.025} & \withsd{0.0511}{0.0038} & \withsd{0.2306}{0.0140} \\
Uniform response probabilities & \withsd{254.60}{8.35} & \withsd{\underline{8.16}}{0.20} & \withsd{0.68}{0.028} & \withsd{0.0552}{0.0058} & \withsd{0.2295}{0.0148} \\
Most likely response only & \withsd{243.85}{10.00} & \withsd{7.87}{0.17} & \withsd{0.67}{0.025} & \withsd{0.0526}{0.0044} & \withsd{0.2187}{0.0112} \\
Without eligibility filters & \withsd{\underline{263.20}}{11.12} & \withsd{8.09}{0.21} & \withsd{\underline{0.70}}{0.023} & \withsd{\underline{0.0588}}{0.0039} & \withsd{0.2214}{0.0127} \\
Choose highest utility automatically & \cellcolor{resultgreen}\withsd{\textbf{272.82}}{7.91} & \withsd{7.96}{0.24} & \withsd{0.69}{0.026} & \withsd{0.0567}{0.0051} & \withsd{\underline{0.2391}}{0.0129} \\
\bottomrule
\end{tabularx}
\caption{Component ablations; aggregation follows Table~\ref{tab:main-seen}. The future representations preserve numerical guidance and filters, while probability interventions change them. Higher is better. Highlighting follows Table~\ref{tab:main-seen}.}
\label{tab:mechanisms}
\end{table*}

\begin{table}[t]
\centering\footnotesize
\setlength{\tabcolsep}{2pt}
\begin{tabularx}{\columnwidth}{@{}>{\raggedright\arraybackslash}X ccc@{}}
\toprule
Predictor & Magentic & SOTOPIA & AgentSense \\
\midrule
Single future & \withsd{0.201}{0.010} & \withsd{0.178}{0.008} & \withsd{0.164}{0.013} \\
Direct expectation & \withsd{0.171}{0.009} & \withsd{0.134}{0.010} & \withsd{0.158}{0.014} \\
Sampled futures & \withsd{0.145}{0.009} & \withsd{0.151}{0.008} & \withsd{0.169}{0.011} \\
\method{} (SFT) & \withsd{\underline{0.139}}{0.010} & \withsd{\underline{0.133}}{0.008} & \withsd{\underline{0.142}}{0.012} \\
Full \method{} & \cellcolor{resultgreen}\withsd{\textbf{0.127}}{0.007} & \cellcolor{resultgreen}\withsd{\textbf{0.116}}{0.011} & \cellcolor{resultgreen}\withsd{\textbf{0.135}}{0.011} \\
\bottomrule
\end{tabularx}
\caption{Normalized prediction RMSE ($\downarrow$) on shared retained cases (94\%); aggregation follows Table~\ref{tab:main-seen}. Tables~\ref{tab:forecast-cost} and~\ref{tab:mechanism-diagnostics} add cost, validity, rankings and regret. Highlighting follows Table~\ref{tab:main-seen}.}
\label{tab:forecast-quality}
\vspace{-1em}
\end{table}

\paragraph{Reporting.}
Tables~\ref{tab:main-seen}--\ref{tab:forecast-quality} report mean $\pm$ SD across five seeds, each averaging three backbones. All test scenarios and focal roles completed (Table~\ref{tab:eval-counts}). Main panels were inspected during development. Appendix~\ref{app:analyses} gives paired intervals from 10,000 scenario/seed bootstrap resamples and a separate three-seed audit.

\subsection{Main Results}\label{sec:results}

\paragraph{Cross-backbone transfer.}
Table~\ref{tab:main-seen} reports held-out scenario families of the training environments. The evaluation backbones are absent from rollout collection, so this tests transfer to new models within familiar environments. \method{} (SFT + GRPO) achieves the best average performance on six of seven metrics. For instance, its Magentic-NL customer utility is 272.82, against 255.56 for sampled futures. The exception is AgentSense information accuracy, where S3AP F\&A (0.95) and single future (0.91) exceed it (0.88). Sampled futures falls below the actor alone on both AgentSense measures, showing that the benefit of sampled predictions varies across environments.

\paragraph{Joint environment and backbone transfer.}
In Table~\ref{tab:main-unseen}, neither the environments nor the evaluation backbones appear in WM training. Even so, \method{} (SFT + GRPO) achieves the best average performance on seven of eight metrics and reaches mediator intelligence of 4.3321, against 3.4387 for S3AP F\&A. The exception is CPV, where SFT-only is lower (0.2281 vs.\ 0.2450). Overall, the final model performs best on 13 of 15 metrics.

\begin{figure}[!t]
\centering
\begin{subfigure}{\columnwidth}
\includegraphics[width=\linewidth]{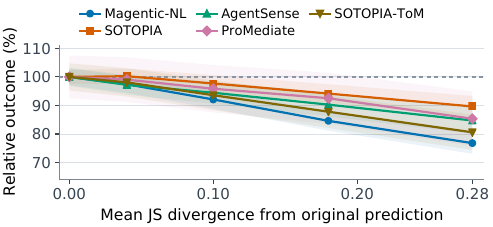}
\caption{Probability perturbation.}\label{fig:probability-sensitivity}
\end{subfigure}
\par
\begin{subfigure}{\columnwidth}
\includegraphics[width=\linewidth]{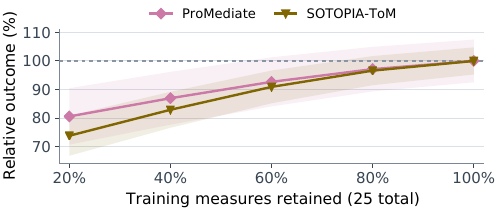}
\caption{Diversity of training outcome measures.}\label{fig:measure-diversity}
\end{subfigure}
\caption{Sensitivity results (\% of full-condition mean), with SD bands. Protocols and raw values: Appendix~\ref{app:sensitivity}.}
\label{fig:sensitivity}
\vspace{-1em}
\end{figure}

\subsection{Discussion}\label{sec:discussion}

\paragraph{Importance of considering alternative futures.}
A social world model should help an agent consider consequences beyond the most likely response. Figure~\ref{fig:intro} illustrates why: predicting acceptance alone favors a higher offer, while considering rejection changes the preferred action. The ablations show a corresponding pattern across the five environments (Table~\ref{tab:mechanisms}). Retaining only the most likely response lowers every primary metric, including Magentic-NL customer utility from 272.82 to 243.85 and SOTOPIA goal achievement from 8.21 to 7.87. Since this intervention changes numerical guidance and eligibility, these results support retaining alternative futures when evaluating and filtering candidate actions.

\paragraph{Importance of probability assignments.}
Beyond representing alternatives, the model must indicate how much each future should influence a decision. Uniform probabilities lower all five primary metrics, although the SOTOPIA change is small (8.21 to 8.16). We further perturb probabilities while holding response texts and conditional outcome values fixed. At pooled average JS divergence 0.28, performance falls to 77--90\% of the original values (Figure~\ref{fig:probability-sensitivity}). Reweighting changes how possible consequences contribute to expected utility and eligibility. These results show that the learned probability assignments carry useful decision information; calibration against observed frequencies remains a separate question.
\vspace{-1mm}
\paragraph{Importance of explaining how outcomes arise.}
Response descriptions further help the actor interpret the predicted consequences. With numerical guidance and filters fixed, full guidance outperforms expected values only in four environments, with paired intervals above zero (Table~\ref{tab:paired-uncertainty}). Adding response text benefits all four, while Magentic-NL remains unchanged under payoff-only filtering. Thus, even when expected values are available, the actor can benefit from descriptions of the responses behind them. Figure~\ref{fig:decision-example} illustrates such a choice. Response clusters also support nonlinear objectives: computing utility within clusters before averaging raises InfoMgmt from 0.2371 to 0.2462 in the separate aggregation comparison (Table~\ref{tab:utility-aggregation}). This remains an approximation and operates within the eligible set (Appendix~\ref{app:mechanism-analysis}). Together, these findings support retaining response information alongside numerical summaries for action selection.
\vspace{-1mm}
\paragraph{Learning diverse social consequences.}
Learning varied consequences improves generalization to unseen environments. As shown in Figure~\ref{fig:measure-diversity}, training DSWM on 25 rather than 5 measures raises consensus change from 0.0506 to 0.0628 and InfoMgmt from 0.1851 to 0.2508, with diminishing gains across intermediate subsets. Shared rollouts and matched supervision volume support attributing these gains to broader outcome supervision. Thus, learning different objectives and costs helps the same model support decisions in unseen environments and settings.

\paragraph{Making multiple-future evaluation practical.}
Compact distributions reduce the cost of evaluating alternative futures. With matched SFT budgets, \method{} (SFT only) exceeds sampled futures on all 15 metrics, though the InfoMgmt gain is only 0.0013 and target aggregation differs. GRPO improves 13 metrics and reduces prediction error, but worsens CPV and leaves information accuracy unchanged (Table~\ref{tab:forecast-quality}; Appendix~\ref{app:mechanism-analysis}). On SOTOPIA, prediction uses 16.2$\times$ fewer tokens and 4.8$\times$ lower latency than 30 sampled futures, including retries; direct expectation remains cheapest (Table~\ref{tab:forecast-cost}). Thus, repeated training simulations support one distributional prediction per action at inference.

\section{Related Work}\label{sec:related}
\vspace{-1mm}
\paragraph{Social world models and dialogue planning.}
Benchmarks cover goal-driven interaction \citep{sotopia,agentsense,promediate,sotopiatom} and cultural pressure \citep{liveculture}. \citet{socialwm} define probabilistic transitions but use one next-state prediction in Foresee-and-Act. \citet{mentis} factor physical and mental updates without training, while \citet{beliefswm} learn collective-belief transitions. Dialogue planners simulate negotiations \citep{lewis2017deal} or search with prompted user simulators \citep{gdpzero}. \method{} instead learns response probabilities and conditional outcomes from rollouts for a separate actor.

\vspace{-1mm}
\paragraph{Decision making with predicted outcomes.}
Expected-utility theory weights utilities by probability \citep{vonneumann1944}, and Monte Carlo evaluation averages rollout returns \citep{sutton2018}. \method{} estimates conditional mean outcomes within response clusters, whereas probabilistic ensembles model dynamics uncertainty \citep{pets} and distributional reinforcement learning models full return distributions \citep{distributionalrl}. Distributional alignment predicts population answers to fixed survey options \citep{evida}; our alternatives are action-conditioned and expressed as generated text. Video-world-model evaluation also distinguishes outcome plausibility from probability accuracy \citep{pawbench}, a distinction relevant here.

\section{Conclusions}\label{sec:conclusion}

We presented \method{}, a distributional social world model that predicts future states as response alternatives, probabilities and outcome values to guide a separate actor. Across five environments and three unseen backbones, it achieved the best performance on 13 of 15 metrics, outperforming other world modeling baselines. Ablations support retaining multiple futures and their probability weights, while richer actor-facing information improves outcomes in four environments.

\label{mainend}
\FloatBarrier
\section*{Limitations}

\paragraph{Simulated partners.}
All agents are language models. Training rollouts use Ornith-1.5-9B, Qwen3.5-9B and Gemma~4~12B, while evaluation uses Ministral~3~14B, MiniMax~H3 and Claude~Sonnet~5. \method{} models the responses of these language-model policies. Transfer to people remains untested, and per-backbone analysis is needed to establish which models benefit.

\paragraph{Coarse targets.}
Fifty continuations per action give reference probabilities in increments of 0.02, so rare responses can be missed. Per-cluster means also omit within-cluster variance, so nonlinear utilities remain approximate. In addition, judge noise enters the labels. Final-horizon targets come from the ordinary actor policy rather than from repeated model-guided selection.

\section*{Ethical Statements}

\paragraph{Dual use.}
A model that predicts how others will respond to an utterance could help an agent pressure or manipulate people in negotiation or persuasion. This work studies interactions among language-model agents in research environments. Any use with people should therefore involve disclosure that an AI agent is participating, consent, and objectives that account for the interests of the counterpart. Furthermore, the eligibility filters in \method{} act on predicted expected values and are not safety mechanisms.

\paragraph{Data and participants.}
This implementation uses scenarios and character profiles from existing research benchmarks and from Magentic Marketplace data. Some scenarios are derived from existing scripts and stories. Interaction partners are simulated. For the judge-validation audit, one study author manually annotated all 400 interaction transcripts, 100 per judged environment. Human involvement was limited to this author annotation, which provides an internal check and may share the assumptions of the study. Appendix~\ref{app:repro} identifies the code and data licences, our adaptations, and the attribution and redistribution terms retained in the planned release.

\paragraph{Judges and social outcome measures.}
Values of social outcome measures such as goal completion, trust or consensus are produced by LLM judges. They are therefore proxies that can carry cultural and stylistic biases of the judges. Their interpretation is limited to the evaluated simulations.

\paragraph{AI assistance.}
We used Codex and Claude to assist with language editing. The authors reviewed and revised the resulting edits and take responsibility for the final content.

\bibliography{references}
\clearpage
\appendix

\section{Social World Model Interface}
\label{app:spec}

The world model uses the generative interface below with a Qwen3-4B-Instruct-2507 backbone adapted by LoRA. The stored prompts and JSON examples retain the implementation term ``scores'' for the outcome measures and their numerical outcome values defined in Section~\ref{sec:overview}. Literal prompt strings, including ``Forecast unavailable or invalid,'' are preserved; the surrounding prose uses the terminology of the main text. Figure~\ref{fig:app-wm-prompt} shows the prompt template.

\paragraph{Input.} The chat template of the base model contains the setting, the profile of the focal agent, visible history, candidate action, outcome-measure definitions, horizons, continuation policy and output schema. Outcome-measure definitions specify name, range, type and direction. Reference continuations, labels and hidden answer keys are excluded.

\paragraph{Output future states.} The output represents possible future states as response clusters, each with representative response text, a probability and conditional mean outcome values. The optional residual bucket, serialized as \texttt{other}, has probability mass and conditional mean outcome values but no response text. A response segment contains the messages and actions of the other parties until the focal agent next acts or the interaction ends. Outcome values summarize the continuations assigned to a cluster, rather than only its representative text.

At inference, the predictor tests the candidate action in a copy of the environment to determine whether it ends the interaction. If so, it requests $K{=}1$ and a single \texttt{(no response)} outcome with probability 1.

\begin{figure*}[t]
\centering
\footnotesize
\begin{verbatim}
[system]
You are a social world model. You are assigned to one agent (the focal agent) in a multi-party
interaction. Given the focal agent's profile, the interaction history as the focal agent sees
it, and ONE action the focal agent proposes to take next, you predict how the other parties
will respond (everything they say or do until the focal agent's next turn, or until the
interaction ends). You output a probability distribution over distinct plausible responses,
and for each response you predict the requested scores.

[user]
## Setting
{setting}

## Focal agent profile
{profile}

## Interaction history (as seen by the focal agent)
{history}

## Proposed focal agent action
{action}

## Scores to predict (for each response)
{one rendered line per requested score}

Predict each score at these horizons:
- immediate: the score of the interaction right after this response
- final: the expected score at the end of the whole interaction if it continues from this
  response

## Continuation policy for final scores
{continuation}

## Output
Give exactly {K} distinct plausible responses of the other parties, written as they would
appear in the transcript (one line per message, formatted like '[Name] text'). Keep each
possible response concise (at most 60 words); include every requested score and horizon. If
these responses do not cover all outcomes, include an additional top-level "other" object
with "prob" and "scores" for the remaining outcomes; do not invent a representative response
for that bucket. K counts only the explicit responses. Probabilities across responses and the
optional other bucket must sum to 1. Respond with JSON only, in this format:
{"responses": [{"response": "<the other parties' response>", "prob": "<probability>",
  "scores": {"<score name>": {"immediate": "<number>", "final": "<number>"}, ...}}]}
\end{verbatim}
\caption{World-model prompt template, with agent terminology standardized for presentation. Text in braces is filled per example. Only the requested horizons are listed. The rendered JSON format line names every requested outcome measure; it is abbreviated here with placeholders. Line breaks were added for page width.}
\label{fig:app-wm-prompt}
\vspace{-1em}
\end{figure*}

\section{Training Data}
\label{app:data}

\subsection{Collecting Future States}

For each candidate action, we repeatedly restore the same complete environment state and sample response segments from the other parties until the focal agent next acts or the interaction ends. Each restored state retains the same private goals and secrets. Candidate generation and the other parties use temperature 1.0; ordinary focal-agent continuation uses 0.7. Rollout collection uses Ornith-1.5-9B, Qwen3.5-9B and Gemma~4~12B as interaction backbones.

Each candidate receives 50 sampled response segments. The original collection distinguishes sampling a response segment from continuing the interaction to obtain final labels. A single rollout decision at each saved state is shared across all candidate actions and their samples. For a selected state, every sampled branch continues to termination; otherwise, the batch supplies immediate labels only. The supplementary collection continues every branch to termination. Immediate-only batches contribute immediate targets; final targets require complete labels for the requested measure across all 50 samples. This collection rule concerns training labels, independently of whether evaluation episodes complete.

\paragraph{Outcome measurement.} Outcome values come from deterministic calculations where the environment supports them. For example, Magentic-NL deal, revenue and utility follow the Magentic Marketplace analytics \citep{magentic}. All other outcome values come from an LLM judge, using official benchmark rubrics (the seven SOTOPIA-EVAL dimensions, AgentSense goal questions) or additional rubrics (for example partner trust, group cohesion and secret leakage). In the supplementary AgentSense collection, information-question accuracy is computed from the actual multiple-choice answers given by the actor when questions exist. Questions and answer keys are never inserted into WM inputs.

The original-collection labels were produced by an Azure-hosted judge whose raw responses were not retained in auditable form. Their structure can be validated, but a full semantic re-check of every legacy judgment is not possible. The supplementary labels were judged by Azure \texttt{gpt-6-luna} at low reasoning effort. Training labels are never re-graded with the evaluation judge.

\subsection{Constructing Distributional Targets}

Representative actions, identified by structured-act signatures, separate incompatible responses before clustering. Within each group, Qwen3-Embedding-0.6B embeddings are clustered by average linkage at cosine similarity 0.75. The largest $K$ clusters are retained explicitly and the rest form the residual bucket, \texttt{other}. A cluster uses its sampled medoid, or its first member when it has at most two samples. Probabilities are rounded to six decimals. Outcome values are rounded to three decimals, or one for magnitudes of at least 1,000, and then bounded to their ranges.

Probabilities are multiples of $1/50=0.02$; fitting them does not establish held-out calibration. Outcome values share a partition into clusters but lose within-cluster variance and covariance. Medoids retain their sampled wording, including responses longer than the requested 60 words.

\subsection{Query Views}

A candidate batch can support several requests. With seed 0, a random view requests one to four outcome measures, available paraphrases and feasible horizons, adding an unselected official outcome measure with probability 0.6. The cluster count is sampled from $\{1,2,3,4,4,5,6\}$ and capped by available clusters. Deployment-shaped views request decision-time outcome measures, possibly more than four. Supplementary views cover every available outcome measure at both horizons with up to four clusters; recovered views request one previously unqueried final outcome value. A requested measure--horizon pair must have a finite label on every reference sample in the candidate batch. Deployment-shaped views prefer both horizons, then final only, then immediate only, provided all required measures are complete; otherwise, no such view is emitted. Consequently, final conditional means use all members of their clusters, not a subset of completed branches, and probabilities retain the full batch denominator $N$. Missing labels are never imputed.

\subsection{Datasets}

Table~\ref{tab:app-data} counts stored training rows. These are query rows, not independent conversations: different requests for outcome measures can reuse the same context, action and samples.

\paragraph{Unseen environments.} ProMediate and SOTOPIA-ToM contribute no training rows.

\begin{table}[t]
\centering
\small
\begin{tabular}{@{}lrrrr@{}}
\toprule
Data & Mag.-NL & SOTOPIA & AgentS. & Total \\
\midrule
Original & 4,975 & 4,766 & 1,118 & 10,859 \\
Recovered & 1,023 & 0 & 186 & 1,209 \\
New & 43 & 44 & 92 & 179 \\
\midrule
Full pool & 6,041 & 4,810 & 1,396 & 12,247 \\
\bottomrule
\end{tabular}
\caption{Composition of the full training pool. Both training stages use all three sources.}
\label{tab:app-data}
\vspace{-1em}
\end{table}

\paragraph{Recovered final outcome values.} We expose 1,209 previously unqueried final-outcome views: 522 customer-utility and 501 service-revenue views in Magentic-NL, and 186 goal-completion views in AgentSense. This requires no new rollout or judge call. Every reference must have a finite, in-range label, and the batch must not already request that outcome measure at the final horizon. Targets are rebuilt by reclustering the stored samples.

\paragraph{Supplementary collection.} We selected 48 training-split jobs: 12 Magentic-NL, 12 SOTOPIA and 24 AgentSense. The additional AgentSense families were stratified by information-question availability. Each job saves one branch state and proposes four candidates: two ordinary actor proposals, a minimal consequential edit and an alternative trade-off. Invalid or duplicate candidates are dropped. These are action alternatives, not ranked preference pairs.

Each candidate receives 50 continuations, all continued to termination. The 179 retained candidate batches therefore correspond to 8,950 continuation samples.

\paragraph{Splits and filtering.} We respect official splits or hash scenario families into roughly 70/10/20 train/validation/test. Filtering removes non-finite or out-of-range labels and SOTOPIA rows with missing dimensions or an inconsistent overall mean. The prediction splits contain 2,137 validation, 2,731 test, 1,530 held-out-specification validation and 1,833 held-out-specification test rows. These filtered, versioned files are not the untouched original test set.

\paragraph{Held-out prediction inputs.}
Table~\ref{tab:forecast-input-counts} counts the test inputs used for prediction diagnostics. Multiple query rows can request different outcome measures or horizons for the same context and action. We therefore also count distinct visible decision contexts and context--action pairs. A context groups rows with the same environment, scenario, focal role, collection-model assignment, branch index, visible history and continuation policy; an action is counted once within that context. These are visible-context counts, rather than independently verified counts of complete hidden-state snapshots.

The latest diagnostic run retains these query inputs and candidate actions and replaces their stored references with 50 freshly sampled continuations per action. These reference continuations provide the comparison outcomes; they are distinct from the 30 model-generated continuations used by the sampled-futures baseline. The test file contains 926 rows requesting final outcomes. The held-out-specification file reuses the same contexts and actions with different requests, so its rows do not add new decision contexts. Counts in Table~\ref{tab:forecast-input-counts} describe the input pool before repetition across evaluation backbones and seeds or validity-based diagnostic exclusions. The prediction-quality comparisons retain 94\% of diagnostic cases and use the same retained set for all methods (Appendix~\ref{app:diagnostic-handling}).

\begin{table*}[t]
\centering\small
\begin{tabular}{@{}lrrrrr@{}}
\toprule
Environment & Test queries & Final queries & Visible contexts & Context--action pairs & Held-out-spec. queries \\
\midrule
Magentic-NL & 1,316 & 485 & 206 & 936 & 936 \\
SOTOPIA & 1,036 & 335 & 133 & 511 & 518 \\
AgentSense & 379 & 106 & 95 & 379 & 379 \\
\midrule
Total & 2,731 & 926 & 434 & 1,826 & 1,833 \\
\bottomrule
\end{tabular}
\caption{Held-out prediction-input inventory. Final queries are test rows explicitly requesting the final horizon. Query files share the contexts and candidate actions shown here. Latest diagnostic references use 50 continuations per action. These source counts are distinct from the online episode counts in Table~\ref{tab:eval-counts} and from post-exclusion diagnostic denominators.}
\label{tab:forecast-input-counts}
\vspace{-1em}
\end{table*}

\section{Training}
\label{app:training}

We train \method{} from Qwen3-4B-Instruct-2507 on the full pool of 12,247 query rows described in Appendix~\ref{app:data}. SFT first learns the serialized response-cluster targets; GRPO then optimizes the prediction-quality reward in Eq.~\ref{eq:reward}. Tables~\ref{tab:main-seen} and~\ref{tab:main-unseen} report both the SFT checkpoint and the final SFT + GRPO checkpoint. Table~\ref{tab:app-hparams} records the training settings.

\paragraph{Supervised fine-tuning.}
The loss is cross-entropy over assistant completion tokens, including the end-of-turn boundary in the chat template. Prompt and padding tokens are masked. Rows longer than 8,192 tokens are excluded before balancing. Let $m$ be the number of eligible rows in the largest environment. Each environment with $n$ eligible rows is repeated $\lfloor m/n\rfloor$ times in full and topped up with $m \bmod n$ rows sampled without replacement. This schedule is built once and reshuffled for each epoch. If every row is eligible, an epoch has $3\times6{,}041=18{,}123$ presentations. The training pool has no data cap or horizon-specific subsampling; token limits determine the eligible rows. Every 200 steps, the run evaluates on 400 shuffled validation rows. We keep the final checkpoint rather than selecting one on validation loss.

\paragraph{Reinforcement learning.}
A row is eligible if its prompt has at most 6,144 tokens and its target completion at most 4,096. Rows with exact-string reference entropy above one nat are listed twice. At each step, the trainer takes four prompts and samples eight completions for each, at temperature 1.0 with at most 4,096 new tokens. Truncated completions contribute their computed rewards to group statistics but no tokens to the policy loss. For completion $o_i$ with reward $R_i$, the advantage $\hat{A}_i$ is $R_i$ minus the group mean, divided by the group standard deviation. We use the clipped GRPO surrogate with DAPO token normalization \citep{dapo},
\begin{equation}\label{eq:grpo-loss}
\mathcal{L}_{\mathrm{GRPO}}
 = -\frac{1}{T}\sum_{i,t} m_{i,t}\,\ell_{i,t},
\end{equation}
where $m_{i,t}$ masks padding and truncated completions, $T$ counts active completion tokens in the accumulated batch, and
\begin{equation}
\begin{split}
\ell_{i,t}=\min\big\{&\rho_{i,t}\hat A_i,\\
 &\operatorname{clip}(\rho_{i,t},1-\epsilon,1+\epsilon)\hat A_i\big\}.
\end{split}
\end{equation}
Here $\rho_{i,t}$ is the token probability ratio between the current and sampling policies, and $\epsilon=0.2$. A small denominator constant stabilizes advantage normalization when group rewards have zero variance. With $\beta=0$, there is no KL penalty or frozen reference model. These settings use the DAPO loss normalization, not its full training recipe.

To retain full-data coverage, GRPO makes one shuffled pass over the eligible, entropy-weighted query pool. Its step count follows the retained pool size and the four-prompt batch, rather than a fixed step cap. The final batch repeats rows if needed to fill it. The eligible dataset is not subsampled. We keep the final checkpoint. Token-budget exclusions are logged separately for SFT and GRPO, so using the full pool does not imply that every stored row meets the limits of both SFT and GRPO.

\paragraph{Reward.}
The reward uses the $N$ stored reference response segments $y_n$, each with weight $1/N$, and their observed outcome values. Reference segments and predicted response texts are embedded with Qwen3-Embedding-0.6B. A reference segment is assigned to the explicit response with the highest cosine similarity, provided the similarity is at least 0.75 and the two carry the same representative actions; otherwise it is missed. Let $m_k$ be the mass assigned to response $k$, $m_o$ the missed mass, $p_k$ the predicted probabilities and $p_o$ the predicted residual mass. Then
$R_{\mathrm{dist}} = 1-\tfrac12\big(\sum_k|p_k-m_k|+|p_o-m_o|\big)$,
$R_{\mathrm{cov}} = 1-m_o$, and
$R_{\mathrm{prec}} = \sum_{k:\,m_k>0}p_k$.
The reward evaluates only the measures and horizons requested by that query, so an immediate-only query has no final-horizon error term. Complete-label checks occur during query construction, before SFT or GRPO. For each requested outcome measure and horizon, we compute the mean squared range-normalized error between the observed outcome value of each reference and the conditional mean for its assigned response cluster. Missed references are assigned to \texttt{other} when the prediction contains one. Without a residual bucket, unmatched references are excluded from this outcome-prediction term and remain penalized by the coverage and distribution terms. An outcome measure with no assigned references receives error 1, and $R_{\mathrm{score}}$ is one minus the mean error over requested outcome measures and horizons. $R_{\mathrm{div}}$ is one minus the fraction of explicit response pairs whose act-compatible cosine similarity exceeds 0.92, and it equals 1 when the prediction has a single response. The gate $g$ is zero when the parser rejects the prediction. Before reward computation, the parser renormalizes probabilities and clips outcome values, so the reward is computed on the repaired prediction.

\begin{table}[t]
\centering
\small
\begin{tabularx}{\columnwidth}{@{}lX@{}}
\toprule
Setting & Value \\
\midrule
LoRA & rank 64, $\alpha$ 128, dropout 0.05; attention and MLP projections \\
SFT schedule & 2 balanced epochs; final checkpoint \\
SFT learning rate & $10^{-4}$; cosine, 3\% warmup \\
SFT batch / length & 1 $\times$ accumulation 16; 8,192 tokens \\
GRPO schedule & 1 pass over the eligible weighted pool \\
GRPO learning rate & $2\times10^{-6}$; constant after 10 warmup steps \\
GRPO batch & 4 prompts $\times$ 8 completions \\
GRPO token limits & prompt 6,144; completion 4,096 \\
GRPO clip / KL & $\epsilon=0.2$; $\beta=0$ \\
Precision / repetitions & bfloat16 / five training seeds \\
\bottomrule
\end{tabularx}
\caption{Settings for each SFT and SFT + GRPO training run. Main Tables~\ref{tab:main-seen}--\ref{tab:forecast-quality} average five training/evaluation seeds; the separate reliability audit uses three.}
\label{tab:app-hparams}
\vspace{-1em}
\end{table}

\section{Benchmark Implementations}
\label{app:benchmarks}

\paragraph{Domains and goals.} Magentic-NL tests whether anticipating acceptance, counteroffers and rejection helps customers and businesses negotiate useful agreements. SOTOPIA tests coordination, collaboration, information exchange and competition between characters with private goals. Its evaluation includes believability and relationships as well as goal achievement. AgentSense spans family and friendship scenes, schools, workplaces and public interactions; characters pursue individual goals and may hold private information.

ProMediate asks whether predictions learned from direct participation transfer to helping other parties resolve multi-topic disagreements. The focal agent is the mediator. SOTOPIA-ToM spans sectors including healthcare, technology, finance and education. Agents must share useful facts with the appropriate recipients, seek missing information and respect disclosure restrictions. We use it as a transfer environment without adding a theory-of-mind component to \method{}.

Because our collection procedure requires repeated continuations from the same interaction state, we implement all five benchmarks in a shared framework that can save and restore complete environment states. We port scenario data, agent prompts and outcome-measure calculations where they are available. Table~\ref{tab:app-bench} lists the deviations. Because of these deviations, outcome values are not comparable with official leaderboard numbers. Using SOTOPIA-ToM as a benchmark does not mean that \method{} contains a theory-of-mind or belief model.

\begin{table*}[t]
\centering
\footnotesize
\begin{tabularx}{\textwidth}{@{}l l X X r@{}}
\toprule
Benchmark & Use & Runner and deviations & Main metrics (native scale) & Test \\
\midrule
Magentic-NL & train & New two-party customer/service negotiation built from Magentic Marketplace data \citep{magentic} (restaurant and home-contractor scenario families). It is not the official marketplace benchmark. Each turn is free text plus an optional act (offer, counter, accept, decline); at most 12 messages; scenario-defined price cap; deterministic negotiation style per scenario. & customer utility, service revenue, needs met (deterministic) & 90 \\
SOTOPIA & train & 450 benchmark combinations and prompts \citep{sotopia}; at most 20 turns or until an agent leaves. The local judge uses the SOTOPIA-EVAL rubrics in an object schema. & seven SOTOPIA-EVAL dimensions, including goal (0--10) & 70 \\
AgentSense & train & Official template \citep{agentsense}; random next speaker with no repeats; 15 messages; first paragraph only, 128 tokens. Official goal questions, but a single judge instead of the official three-model ensemble. & goal completion (0--1); information questions where available & 270 \\
ProMediate & unseen & Official cases, personas, conflict-mode, mediator and judge prompts and metric formulas \citep{promediate}. Compact motivation-ranked turn-taking replaces the released engine; at most 40 utterances (official: $4\times$parties$\times$topics); known released-code bugs avoided; initial judgments precomputed. & consensus change; topic-level efficiency; mediator intelligence & 18 \\
SOTOPIA-ToM & unseen & We reconstruct the runner and judge from the released gold set and the paper \citep{sotopiatom}: round-robin turns, public and direct-message channels, at most 60 turns, early stop. The DA/IA/CPV/EFF judge prompts are our reimplementation; the formulas follow the paper; metrics are computed for the focal agent. & InfoMgmt; DA, IA, CPV, EFF & 160 \\
\bottomrule
\end{tabularx}
\caption{Benchmarks as implemented in our shared framework. ``Test'' counts held-out scenario configurations, before crossing focal roles, five seeds and three backbones (Table~\ref{tab:eval-counts}).}
\label{tab:app-bench}
\vspace{-1em}
\end{table*}

\paragraph{Agents and judges.} Rollout collection uses Ornith-1.5-9B, Qwen3.5-9B and Gemma~4~12B. Evaluation uses the disjoint set of Ministral~3~14B, MiniMax~H3 and Claude~Sonnet~5 in both training and unseen environments. The WM backbone remains Qwen3-4B-Instruct-2507; these are changes to the interaction backbones, not the predictor. Evaluation enumerates every eligible focal role in each test scenario; ProMediate evaluates only the mediator. The evaluation judge is GPT-5.6-Sol, and training labels are not re-graded with it. Cache keys distinguish judge configurations; earlier verdicts are not reused. The five runs vary training and evaluation seeds; scenario and model assignments are matched across methods within each run.

\paragraph{Evaluation coverage.} We use every test-split scenario and keep validation separate. Table~\ref{tab:eval-counts} counts scenarios and eligible focal-role configurations directly from the benchmark loaders and split function. The online episode count per applicable method is $3\times5\times\sum_{s\in\mathrm{test}}|\mathcal{R}_s|$, where $\mathcal{R}_s$ is the set of eligible focal roles in scenario $s$. Thus, the counts cross three backbones and five seeds; they are not obtained by subtracting prediction-query rows from scenario counts. Every scheduled episode completed, and the same grid is used for the online component ablations. GDP-Zero is evaluated only in Magentic-NL and SOTOPIA.

\begin{table*}[t]
\centering\small
\setlength{\tabcolsep}{5pt}
\begin{tabularx}{\textwidth}{@{}X rrrrrr@{}}
\toprule
Environment & Total & Train & Val. & Test & Test role configs. & Episodes / method \\
\midrule
Magentic-NL & 439 & 306 & 43 & 90 & 180 & 2,700 \\
SOTOPIA & 450 & 325 & 55 & 70 & 140 & 2,100 \\
AgentSense & 1,225 & 865 & 90 & 270 & 650 & 9,750 \\
ProMediate & 18 & 0 & 0 & 18 & 18 & 270 \\
SOTOPIA-ToM & 160 & 0 & 0 & 160 & 536 & 8,040 \\
\bottomrule
\end{tabularx}
\caption{Scenario inventory and complete online evaluation grid. Total, Train, Val. and Test count scenario configurations; Train describes the eligible scenario split, not the number of prediction-query rows. Test role configurations count each eligible focal role separately, per backbone and seed. Episodes multiplies this count by three backbones and five seeds. Counts apply to Tables~\ref{tab:main-seen}--\ref{tab:mechanisms} for each applicable method, not to the separate reliability audit or saved-state prediction diagnostics.}
\label{tab:eval-counts}
\vspace{-1em}
\end{table*}

Magentic-NL contributes 90 customer-focal and 90 service-focal configurations, giving 1,350 episodes for the payoff column for each role. SOTOPIA evaluates both characters in each of 70 configurations. AgentSense contributes 650 character-focal configurations across 270 scenarios, while SOTOPIA-ToM contributes 536 across 160 scenarios. ProMediate has six cases, each in three conflict modes, with the mediator focal. The test splits contain 63 Magentic-NL customer families, 14 SOTOPIA environment families and 54 AgentSense template families; scenario variants within a family are not independent families.

For AgentSense information accuracy, 183 of the 650 test role configurations contain information questions, with 352 questions in total. Across three backbones and five seeds, this gives 2,745 evaluated episodes containing 5,280 question answers per method. We first average correctness over the available questions within each episode, then average the episode-level accuracies; episodes without such questions do not enter this metric.

Within each backbone and seed, online metrics average the relevant episode subset. We then average the three backbone means with equal weight and report mean and sample SD across the five seed-level means. Main test scenarios were inspected during development. This inspection did not lead to changes in training choices, prompts, filters or hyperparameters. The test results therefore did not guide revisions to these settings. The separate reliability audit in Appendix~\ref{app:analyses} retains its three-seed, 300-pair protocol; its denominators and intervals do not describe the main tables.

\begin{table*}[t]
\centering
\footnotesize
\begin{tabularx}{\textwidth}{@{}p{.22\textwidth} X p{.17\textwidth}@{}}
\toprule
Method & Information at decision time & WM training \\
\midrule
Actor alone & Ordinary action prompt, without candidate selection & none \\
Candidates only & Up to four actor proposals; selection without predictions & none \\
Single future & One predicted response segment and its outcome values & matched SFT \\
Direct expectation & Expected value of each requested outcome measure and horizon; no response text & matched SFT \\
Sampled futures & 30 independent response segments with outcome values per candidate & matched SFT \\
S3AP F\&A & A prompted next-state prediction and the Foresee-and-Act procedure \citep{socialwm} & none \\
GDP-Zero & Open-loop tree search with prompted simulation of both sides \citep{gdpzero}; two-party settings only & none \\
\method{} (SFT only) & Response-cluster probabilities, descriptions and conditional mean outcome values & matched SFT \\
\method{} (SFT + GRPO) & The same interface after reinforcement learning & SFT + GRPO \\
\bottomrule
\end{tabularx}
\caption{Evaluated decision-time methods. Matched SFT uses the same base model, LoRA settings, source rollout pool and SFT budget. It does not include the additional GRPO stage.}
\label{tab:app-baselines}
\vspace{-1em}
\end{table*}

\section{Applying DSWM at Inference Time}
\label{app:selection}

At each decision point of the focal agent, the model-guided policy runs seven steps.

\paragraph{1. Candidates.} The actor samples up to $M{=}4$ actions at temperature 1.0 with its ordinary action prompt. Each action is converted to the standard format required by the environment; for example, \texttt{leave} and \texttt{none} in SOTOPIA carry no spoken payload. Invalid actions are dropped. Duplicates are removed by speaker, text, structured act, channel and recipients. If no valid action remains, the policy returns the same fail-closed no-op as the actor-alone baseline.

\paragraph{2. Predictions.} Each candidate gets its own prediction request, with the outcome measures listed below, both horizons and $K{=}4$, or $K{=}1$ for an action that ends the interaction (Appendix~\ref{app:spec}). Requests run concurrently on vLLM with LoRA adapters, using temperature 0.7, top-$p$ 0.8, top-$k$ 20, at most 4,096 new tokens, and a deterministic seed within each evaluation run.

The parser checks that required fields, outcome values, horizons and response counts are present. It normalizes probabilities to sum to one and clips outcome values to their ranges. Missing or non-finite values make a prediction invalid. The served predictor makes one generation attempt per candidate; parse failures are not retried. Because of normalization and clipping, a valid parse does not show that the raw output was calibrated or within range.

\paragraph{Outcome measures at inference.} SOTOPIA requests all seven native dimensions with equal weights. AgentSense requests goal completion, estimated information reasoning and secret leakage (minimized, weight 0.25). ProMediate requests final consensus, consensus change, topic-level efficiency, mediator intelligence and tension (minimized, weight 0.25). SOTOPIA-ToM requests DA, IA, CPV and EFF; CPV is minimized and is the only active constraint among these five environments. Other unspecified weights are 1. Magentic-NL requests role-specific payoff, deal reached and needs met; only payoff has positive weight. The reasoning estimate for AgentSense is a planning proxy; reported information accuracy comes from actual question answering.

\paragraph{3. Expected outcome values.} For each candidate, the policy computes the probability-weighted mean of every outcome measure at both horizons, including the \texttt{other} bucket.

\paragraph{4. Filters.} Two eligibility filters run on the normalized final expected outcome values.
\begin{itemize}[leftmargin=*,itemsep=1pt,topsep=2pt]
\item \textbf{Constraints.} For each outcome measure marked as a constraint, a candidate stays eligible only if its expected constraint cost is within 0.02 of the lowest constraint cost in the current eligible set. Constraints are applied sequentially in measure order. This is a prediction-based safeguard, not a safety guarantee.
\item \textbf{Dominance.} A candidate is removed if another candidate is at least as good on every objective and constraint measure and strictly better on at least one.
\end{itemize}
Only positive-weight outcome measures and active constraints enter the dominance comparison; in Magentic-NL, this is payoff alone. Candidates with invalid predictions are not eligible. If every prediction is invalid, all candidates become allowed, no utility line is shown, and the fallback is logged.

\paragraph{5. Combined utility.} For each cluster, every final outcome value is mapped to $[0,1]$, flipped if it is minimized, clipped, and combined with the measure weights. SOTOPIA-ToM uses a weighted geometric mean; all other environments use a weighted arithmetic mean. The result is averaged over clusters using their probabilities. For a nonlinear combination, this is an approximation, because variation inside a cluster is not represented. The actor sees utilities for eligible candidates on an additional line, but is not required to maximize them.

\paragraph{6. Selection.} The same actor backbone receives the selection prompt in Figure~\ref{fig:app-select-prompt}. For each candidate it shows:
\begin{itemize}[leftmargin=*,itemsep=1pt,topsep=2pt]
\item the action;
\item its immediate and final expected outcome values;
\item its two most probable response descriptions with their probabilities, cut to 60 words;
\item the probability mass of the \texttt{other} bucket.
\end{itemize}
After the candidates, the prompt lists the eligible action IDs and the combined utility line. A JSON schema restricts the output to an allowed ID. Decoding uses temperature 0, at most 64 tokens and a fixed seed. If two attempts fail, the episode fails explicitly. The policy never silently substitutes another action. For the separate qualitative case in Figure~\ref{fig:decision-example}, we also retain the reasoning produced by the actor separately from its final candidate ID. The selection prompt is unchanged; the final ID determines the executed action, while the reasoning is shown for interpretation.

\paragraph{7. Execution.} The environment executes a copy of the selected candidate exactly. The actor does not rewrite it.

\paragraph{Logging and the candidates-only baseline.} Each decision logs the candidates, predictions, eligible set, raw actor selection output, finish reason, fallback flag and outcome-measure specifications. The candidates-only baseline runs steps 1, 6 and 7 with the candidate list alone and all IDs allowed.

\paragraph{Continuation mismatch.} Final predictions are values under ordinary continuation. Online, however, model-guided selection is applied again at every later focal agent turn. The targets do not model this repeated use, and the supplementary collection does not remove the mismatch.

\begin{figure*}[t]
\centering
\footnotesize
\begin{verbatim}
[system]
Select one of the supplied actions for the focal agent. Return only the required candidate_id
JSON; do not generate a new action.

[user]
Setting:
{setting}

Your profile:
{profile}

Visible history:
{history}

[user]
## Private action selection
Scores for you (predictions, not observed facts):
{one rendered line per decision score}

Candidate 1: {action}
immediate: {score}={expected value, 3 decimals}; ...
final: {score}={expected value, 3 decimals}; ...
Possible response ({probability}%): {most probable predicted response, first 60 words}
Possible response ({probability}%): {second most probable predicted response}
Other responses: {probability}%
...
Allowed candidate IDs: [{eligible IDs}].
Combined expected utility (higher is better): {ID}: {value, 4 decimals}, ...
Select a candidate that advances your goals while respecting the stated constraints. Predicted
responses are hypothetical; do not assume they actually occurred. Return ONLY a JSON object
{"candidate_id": <integer>}. Do not write dialogue or reasoning here. The selected candidate
will be executed exactly.
\end{verbatim}
\caption{Selection prompt, with agent terminology standardized for presentation. If the prediction for a candidate is invalid, its prediction lines are replaced by ``Forecast unavailable or invalid; do not treat it as evidence.'' The utility line is omitted in the candidates-only baseline and in the all-invalid fallback. The candidates-only baseline lists actions only. Line breaks were added for page width.}
\label{fig:app-select-prompt}
\vspace{-1em}
\end{figure*}

\section{Baselines and Variants}
\label{app:baselines}

Table~\ref{tab:app-baselines} lists the evaluated methods. Within each panel, methods share the focal actor, partner policies, scenarios, judges and episode seeds. Trained prediction baselines also share the candidate-generation and selection procedure. For S3AP Foresee-and-Act and GDP-Zero, we follow the implementations released by their authors exactly rather than redesigning their planning procedures. We connect these methods to the shared benchmark interfaces and evaluation backbones described in Appendix~\ref{app:benchmarks}; GDP-Zero is used only in two-party settings. Thus, these comparisons concern complete systems. Table~\ref{tab:external-baseline-config} records their planning and generation settings. In each evaluation run, the internal world model in S3AP uses the same backbone as its actor; GDP-Zero shares that backbone across its actor, partner simulator and planner. Both therefore use Ministral~3~14B, MiniMax~H3 and Claude~Sonnet~5 across the three backbone conditions, rather than the older models in the released example commands.

\paragraph{Released baseline settings.} S3AP Foresee-and-Act generates a provisional action, predicts one next social state including the mental states of the participants, and refines the action once. The released world-model and refinement calls request temperature 0.7. GDP-Zero uses the experimental runner configuration: 20 MCTS simulations, three realizations per open-loop state, exploration coefficient $c_{\mathrm{puct}}=1$ and initial value $Q_0=0$. The interactive demo instead uses 10 simulations and $Q_0=0.25$. Its actor, simulator, action-prior and value-query temperatures retain the released defaults. These are inference-time baselines; we do not fine-tune their models.

\begin{table*}[t]
\centering\small
\begin{tabularx}{\textwidth}{@{}l l X@{}}
\toprule
Method & Component & Setting \\
\midrule
S3AP F\&A & Planning & One provisional action, one next-state prediction, one refinement \\
 & Prediction / refinement & Temperature 0.7 / 0.7; structured output \\
 & Models & Internal world model matches the evaluation backbone of the focal actor \\
\midrule
GDP-Zero & Search & 20 simulations; 3 realizations per state; $c_{\mathrm{puct}}=1$; $Q_0=0$ \\
 & Simulated utterances & Actor temperature 0.7; partner temperature 1.1; 128-token limit; 5-turn history window \\
 & Action prior & 15 samples at temperature 1.0; additive smoothing 1.0 \\
 & Value estimate & 10 samples at temperature 1.1 \\
 & Models & One evaluation backbone shared by actor, partner simulator and planner \\
\bottomrule
\end{tabularx}
\caption{External-baseline configurations. GDP-Zero uses the 20-simulation, zero-initial-value setting from the experimental runner. Other algorithmic settings follow the released implementations; benchmark interfaces, evaluation backbones and judge are shared as described in Appendix~\ref{app:benchmarks}.}
\label{tab:external-baseline-config}
\vspace{-1em}
\end{table*}

Parameter definitions are traceable to the released source: the \texttt{SocialWorldModelAgent} and \texttt{SocialWorldModel} implementations in S3AP,\footnote{Source inspected at \href{https://github.com/XuhuiZhou/social-world-model/tree/b9e0766486b2d535cde841872ab3fc952d68df06}{\texttt{XuhuiZhou/social-world-model}, \texttt{b9e0766}}.} and the \texttt{runners/gdpzero.py}, \texttt{core/players.py} and \texttt{core/mcts.py} files in GDP-Zero.\footnote{Source inspected at \href{https://github.com/jasonyux/GDPZero/tree/e9f10e089953205a38b2e44857089f96f656c20e}{\texttt{jasonyux/GDPZero}, \texttt{e9f10e0}}.} These links identify the source inspected for the parameter descriptions; the exact historical checkout identifiers of the evaluation runs were not retained in this report.

\paragraph{Matched training.} Prediction baselines use the same base model, LoRA settings and source rollout pool, with matched SFT presentations and optimizer steps. Each trained prediction baseline is retrained for each of the five main-experiment seeds. They do not use GRPO. Comparing them with \method{} (SFT only) keeps the training stage matched, while comparing with the final model also reflects post-training. The within-model variants in Table~\ref{tab:mechanisms} instead modify predictions or selection from one fixed final checkpoint.

\paragraph{Targets and aggregation.} Single future and sampled futures both learn individual response segments with their observed outcome values. Their decision-time predictions contain one and 30 independently generated segments, respectively. Direct expectation learns only means of the raw rollout outcome values, with no response text or separate combined-utility target. The shared action-selection procedure derives expected outcome values from each format and applies the same eligibility rules. In SOTOPIA-ToM, direct expectation computes utility from predicted means, and sampled futures computes it from the mean values across samples. By contrast, \method{} averages utilities computed from cluster means. Appendix~\ref{app:analyses} examines this aggregation difference separately. Matching SFT steps does not equalize target lengths or generation cost. Checkpoint identities and exact per-query sample selection for these two generative baselines are not specified here.

\paragraph{Saved states and online episodes.} Prediction-quality comparisons give every predictor the same saved states and candidate sets. Online histories can diverge after the first different choice, so these comparisons are paired at the episode level, not by identical candidates at every later turn. Table~\ref{tab:forecast-cost} reports prediction accuracy and inference cost on the saved-state SOTOPIA panel; it does not measure total training cost or end-to-end episode latency.

\section{Metrics and Analysis Protocol}
\label{app:metrics}

Tables~\ref{tab:main-seen} and~\ref{tab:main-unseen} report benchmark outcome values on their native scales, and Tables~\ref{tab:mechanisms} and~\ref{tab:forecast-quality} examine components and prediction quality. We do not average incompatible measures across benchmarks.

\paragraph{Evaluation panels.} Tables~\ref{tab:main-seen}--\ref{tab:forecast-quality} report means and SDs across five seeds after averaging the three evaluation backbones within each seed. The reliability audit reports paired differences and scenario-cluster bootstrap intervals on a separate evaluation schedule with three training seeds (Appendix~\ref{app:analyses}). Those intervals cannot be assigned to the main-table means. Every scheduled episode completed for every evaluated method, environment and focal role, with zero episode failures and no failure-based exclusions. This applies to the main evaluations and the separate reliability audit. Table~\ref{tab:eval-counts} gives the main online denominators, including the Magentic-NL role subsets and AgentSense question-bearing subset described alongside it. Several reported metrics overlap: SOTOPIA overall includes goal, and InfoMgmt combines DA, IA, CPV and EFF. Counts of best means therefore do not represent independent statistical tests.

\paragraph{Reported scales and directions.} For the main outcome metrics in Tables~\ref{tab:main-seen} and~\ref{tab:main-unseen}, higher is better except CPV. Magentic-NL utility and revenue use scenario currency units; needs met is binary. Both payoffs are logged in every episode, but their table columns report the subset where that role is focal, rather than all episodes pooled together. SOTOPIA goal ranges from 0 to 10; overall is the unnormalized mean of seven dimensions, ranging from $-30/7$ to $40/7$. AgentSense goal completion and actual information-question accuracy lie in $[0,1]$.

ProMediate consensus change is mean consensus over the closing ten utterances minus the opening ten; topic efficiency averages per-topic improvement per relevant utterance. Both range from $-1$ to 1. Mediator intelligence ranges from 1 to 5. SOTOPIA-ToM reports disclosure alignment (DA), inquiry alignment (IA), critical privacy violation rate (CPV), and information acquisition efficiency (EFF), each in $[0,1]$. InfoMgmt is computed for the focal agent within each episode as $(\mathrm{DA}\,\mathrm{IA}\,(1-\mathrm{CPV})\,\mathrm{EFF})^{1/4}$ and then averaged across episodes. Consequently, it cannot be recovered by applying this formula to the means in the other columns.

\paragraph{Prediction diagnostics.} Table~\ref{tab:mechanism-diagnostics} reports normalized error in expected outcome values, candidate-ranking agreement and reference-estimated regret in Magentic-NL, SOTOPIA and AgentSense. On shared saved SOTOPIA states, Table~\ref{tab:forecast-cost} additionally reports raw request validity, generated tokens and prediction latency per candidate. The sampled-futures validity check covers the complete 30-sample request. Raw validity is assessed before parser normalization or clipping. Accuracy metrics include repaired predictions and use the common retained 94\% of cases. Sampled futures retries until it obtains 30 usable predictions per candidate; token and latency figures include retries. Appendix~\ref{app:diagnostic-handling} details these rules, and Appendix~\ref{app:analyses} describes the separate reliability audit.

\paragraph{Calibration.} Probability calibration is not evaluated here. A calibration study would require independently defined observable events, such as structured acts or termination, and fresh reference continuations. Such events provide shared categories across differently worded response clusters. Agreement with the training matcher, low outcome-value error and judge--human correlation do not establish calibrated response probabilities.

\section{Reproducibility Details}
\label{app:repro}

\paragraph{Data and implementation release.} We will publicly release the training data and implementation upon acceptance.

\paragraph{Models.} The WM is Qwen3-4B-Instruct-2507; clustering uses Qwen3-Embedding-0.6B. Actor and judge identities are listed in Appendix~\ref{app:benchmarks}; supplementary labels use Azure \texttt{gpt-6-luna} at low reasoning effort.

\paragraph{Software.} The current environment uses PyTorch 2.11.0 (CUDA 12.9), Transformers 5.5.0 and PEFT 0.19.1. The exact TRL version is not recorded here. vLLM serves predictions with LoRA; its trajectories are stored separately because sampling can differ from Hugging Face generation.

\paragraph{Hardware and compute budget.}
The experimental workflow used eight NVIDIA A100 GPUs with 80\,GB memory each for approximately 80 hours, corresponding to 640 allocated GPU-hours. This budget covers data collection, training and evaluation. It accounts for the local GPU allocation; server-side compute used by externally hosted model APIs is not included in this GPU-hour total.

\paragraph{Resource licences.}
Magentic-NL uses scenario data from the \href{https://github.com/microsoft/multi-agent-marketplace/blob/main/LICENSE}{MIT-licensed Magentic Marketplace repository}. The \href{https://github.com/ljcleo/agent_sense/blob/main/LICENSE}{AgentSense} and \href{https://github.com/microsoft/ProMediate-Eval/blob/main/LICENSE}{ProMediate} repositories also use the MIT licence. SOTOPIA has separate terms for its \href{https://github.com/sotopia-lab/sotopia/blob/main/LICENSE}{MIT-licensed code} and \href{https://huggingface.co/datasets/cmu-lti/sotopia/blob/main/README.md}{CC BY-SA 4.0 dataset}. The \href{https://huggingface.co/datasets/yashwanthys/sotopia-tom}{SOTOPIA-ToM gold dataset} is released under CC BY 4.0.

We use these resources for simulated research interactions, cite their creators, and identify our adaptations in Table~\ref{tab:app-bench}. The local third-party code retains the upstream licence notices. Our release will preserve the applicable notices, source attribution and descriptions of changes. Adapted SOTOPIA data will retain \href{https://creativecommons.org/licenses/by-sa/4.0/}{CC BY-SA 4.0} terms; SOTOPIA-ToM derivatives will retain the attribution and change notices required by \href{https://creativecommons.org/licenses/by/4.0/}{CC BY 4.0}. Repository licences do not independently establish rights to underlying film or television scripts; our release will identify such source material separately rather than apply a blanket licence to all benchmark-derived content.

\paragraph{Seeds and provenance.} Data-query construction uses seed 0. Main Tables~\ref{tab:main-seen}--\ref{tab:forecast-quality} average five training/evaluation seeds; the separate reliability audit uses three training seeds. Seed identifiers and their exact derivation are not tabulated here. Manifests record data and model hashes, software, sampled rows and validation queries. ProMediate requires precomputed initial judgments.

\section{Implementation Settings}\label{app:method-details}

Section~\ref{sec:method} defines the task, targets and reward. This appendix records implementation settings; Appendices~\ref{app:spec}, \ref{app:data} and~\ref{app:selection} give the interface, data and inference-time action-selection protocol.

\subsection{Collection Settings}\label{app:method-collection}

The actor proposes candidate actions at temperature 1.0, and the other parties respond at temperature 1.0. We sample 50 continuations per candidate in both collections, using Ornith-1.5-9B, Qwen3.5-9B and Gemma~4~12B as the interaction backbones. For final labels, the ordinary continuation policy uses actor temperature 0.7 and other-party temperature 1.0. Each training prompt states this policy. In the original collection, one rollout decision determines whether all candidates and samples at a saved state continue to termination or supply immediate labels only. The supplementary collection always continues every branch to termination. Appendix~\ref{app:data} describes the complete-label requirement for requesting final values.

\subsection{Clustering Settings}\label{app:method-modes}

Within each partition defined by representative actions, texts are embedded with Qwen3-Embedding-0.6B and clustered by average-linkage agglomerative clustering on cosine similarity with a threshold of 0.75. The $K$ largest clusters are retained explicitly. Each is represented by its sampled medoid, or its first member for clusters of one or two. Remaining clusters form the residual bucket. Cluster probabilities and outcome-value targets are defined in Eq.~\ref{eq:targets}; rounding details appear in Appendix~\ref{app:data}.

\subsection{Model and Training Settings}\label{app:method-training}

We adapt Qwen3-4B-Instruct-2507 \citep{qwen3} with LoRA \citep{lora}, using rank 64 and $\alpha{=}128$ on attention and MLP projections. SFT masks prompt and padding tokens. Rows longer than 8,192 tokens are excluded, and rows from smaller environments are repeated and sampled to match Magentic-NL. We train for two epochs and keep the final checkpoint; Tables~\ref{tab:main-seen}--\ref{tab:forecast-quality} average over five training/evaluation seeds.

The GRPO matcher requires cosine similarity of at least 0.75 and agreement of representative actions. The GRPO reward and its components are defined in Eq.~\ref{eq:reward}. GRPO samples eight prediction completions per prompt, which are distinct from the 50 environment continuations used as references. It standardizes rewards within each group and uses neither a KL penalty nor a reference model. SFT and GRPO draw on the same 12,247-row pool under their respective token limits, and GRPO lists rows with reference entropy above one nat twice. Appendix~\ref{app:training} gives the remaining optimizer settings.

\begin{figure}[t]
\centering
\includegraphics[width=\columnwidth]{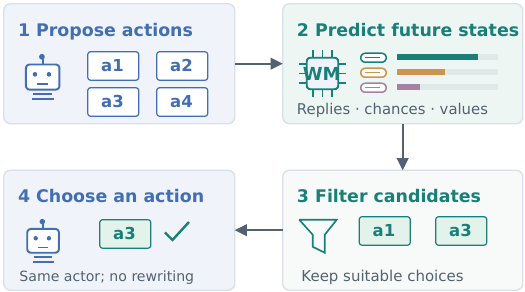}
\caption{Model-guided action selection. Predictions guide filtering and inform action selection by the same actor; the chosen action is executed unchanged. Candidate IDs are illustrative.}
\label{fig:selection}
\vspace{-1em}
\end{figure}

\section{Additional Analyses and Evaluation Protocols}\label{app:analyses}

This appendix gives the protocols for the training, component and prediction diagnostics in Section~\ref{sec:discussion}, followed by utility-aggregation and reliability analyses. Main results, component ablations and prediction diagnostics summarize three evaluation backbones and five training/evaluation seeds. The reliability audit uses a separate three-seed panel, so its intervals and denominators do not apply to the main tables.

\paragraph{Training comparisons.}
The \method{} (SFT only) rows in Tables~\ref{tab:main-seen} and~\ref{tab:main-unseen} use the same base model, source rollouts and matched SFT budget as the three prediction baselines. Comparing the two \method{} checkpoints with the same decision procedure measures the effect of added GRPO for this representation. It does not establish how each baseline would perform with its own GRPO objective. In SOTOPIA-ToM, prediction formats also differ in utility aggregation; Table~\ref{tab:utility-aggregation} therefore varies aggregation within each format.

\paragraph{Component ablations.}
Table~\ref{tab:mechanisms} holds the SFT + GRPO checkpoint fixed within each seed and changes one component per row. Each seed-level result averages the three evaluation backbones; the table reports the mean and SD across five seeds. Removing response text preserves probabilities, expected outcome values, eligibility and utilities. Uniform probabilities are assigned across all prediction buckets, including any residual bucket, before recomputing expected outcome values, utilities and eligibility. The \emph{Most likely response only} variant retains the highest-probability explicit response and sets its probability to one. Removing eligibility filters permits every candidate with a valid prediction. The \emph{Choose highest utility automatically} variant selects the highest-utility eligible candidate, with candidate ID breaking ties. All variants retain the same candidate-generation procedure and all-invalid fallback. Online histories can diverge after different selections; the full \method{} row gives the reference for these variants and matches the corresponding main-table means. All scheduled ablation episodes completed; their counts follow Table~\ref{tab:eval-counts}.

\paragraph{Utility aggregation.}
Table~\ref{tab:utility-aggregation} compares two orders of operation while retaining the checkpoint, eligibility filters based on expected outcome values, candidate-generation procedure and actor prompt for each predictor. Average first computes SOTOPIA-ToM utility from expected outcome values. Combine first computes utility for each of 30 samples and then averages, or computes utility from the conditional means of each \method{} cluster and averages by cluster probability, including any residual bucket. Direct expectation supports only average first. The two combine-first estimators differ because clustering still loses within-cluster variation. Only the utility line changes within each paired comparison; cross-format comparisons retain their training differences. This is a separate aggregation comparison, so its values should be read against its corresponding average-first results.

For sampled futures, combining first raises InfoMgmt from 0.2114 to 0.2238 but also raises CPV slightly, from 0.3097 to 0.3121. For \method{}, InfoMgmt rises from 0.2371 to 0.2462 and CPV falls from 0.2674 to 0.2586. Thus, aggregation order affects decisions, although its effects need not agree across outcome measures.

\begin{table*}[tp]
\centering
\small
\begin{tabularx}{\textwidth}{@{}X l cc@{}}
\toprule
Prediction & Order & InfoMgmt $\uparrow$ & CPV $\downarrow$ \\
\midrule
Direct expectation & Average first & 0.1846 & 0.3382 \\
Sampled futures (30) & Average first & 0.2114 & 0.3097 \\
Sampled futures (30) & Combine first & 0.2238 & 0.3121 \\
\method{} & Average first & \underline{0.2371} & \underline{0.2674} \\
\method{} & Combine first & \cellcolor{resultgreen}\textbf{0.2462} & \cellcolor{resultgreen}\textbf{0.2586} \\
\bottomrule
\end{tabularx}
\caption{SOTOPIA-ToM utility-aggregation comparison, separate from the five-seed main tables; \method{} uses SFT + GRPO. Average first computes utility from expected outcome values; combine first averages utilities of individual samples or cluster means. Only the utility line changes within each prediction format. InfoMgmt is maximized; CPV is minimized. Highlighting follows Table~\ref{tab:main-seen}.}
\label{tab:utility-aggregation}
\vspace{-1em}
\end{table*}

\paragraph{Prediction quality and cost.}
Table~\ref{tab:forecast-cost} evaluates final-horizon SOTOPIA predictions on shared saved states and candidate actions, using only the input visible to the focal agent, and averages the diagnostics across the three evaluation backbones and five training/evaluation seeds. Each candidate action has 50 fresh reference continuations under the ordinary continuation policy. Table~\ref{tab:forecast-input-counts} gives the held-out input counts, separately from the online evaluation grid. RMSE measures errors in expected outcome values after range normalization. Rank is within-state Kendall's $\tau_b$ between predicted and reference candidate values under a common objective. Raw validity is the percentage of complete requests passing schema, finite-value, range, response-count and probability-sum checks before normalization or clipping; a sampled-futures request includes all 30 samples. Generated tokens sum all model outputs per candidate, including retries. Latency covers the prediction request and its retries under the shared serving setup, not the full agent episode.

\paragraph{Prediction repairs and shared exclusions.}\label{app:diagnostic-handling}
The prediction-quality comparisons in Tables~\ref{tab:forecast-quality}, \ref{tab:forecast-cost} and~\ref{tab:mechanism-diagnostics} include predictions that become usable after parser repair. Unrecoverable predictions are excluded. For sampled futures, we retry failed generations until each candidate has 30 usable predictions, then average their outcome values. Both token counts and latency include these retries. After repair, retries and exclusion, all methods are evaluated on the same retained cases, which account for 94\% of the diagnostic cases. RMSE, ranking agreement and regret therefore compare methods on a shared subset, rather than separate method-specific subsets. These accuracy results are conditional on retention; raw validity is measured before repair and is distinct from the shared retention rate. The 6\% diagnostic exclusion rate concerns prediction-quality evaluation, whereas the online episode completion counts follow Table~\ref{tab:eval-counts}.

The SOTOPIA cost diagnostics describe this environment and horizon only. Integer retained counts by environment, the treatment of ranking ties, hardware, batching and repeated timing variability are not included in the current tables. The diagnostics therefore do not support a general hardware-independent speedup claim. Probability calibration also requires a separate evaluation with shared observable event categories and fresh reference continuations.

\paragraph{Reliability and judge validation.}
Table~\ref{tab:reliability} compares \method{} (SFT + GRPO) with 30 sampled futures on separate paired evaluation panels, using three WM training seeds and 300 scheduled paired episodes per environment. All 300 pairs completed for both methods in every environment, with zero episode failures. Pairs therefore equals the scheduled count, and no episode is excluded for failure. The difference is \method{} minus sampled futures. Its primary measure is customer utility for Magentic-NL, goal achievement for SOTOPIA and AgentSense, consensus change for ProMediate, and InfoMgmt for SOTOPIA-ToM. Scenario-cluster bootstrap intervals keep repeats and roles from the same scenario together. Each failure rate uses all 300 scheduled episodes for its own method. This audit is separate from the five-seed main grid; scenario-family disjointness from that exhaustive test grid is not established here.

The intervals exclude zero for Magentic-NL, SOTOPIA, AgentSense and SOTOPIA-ToM, but include zero for ProMediate. Therefore, this audit supports improvements on four primary measures, while the ProMediate difference remains inconclusive. Per-seed estimates and seed identifiers are not tabulated, so the pooled intervals should not be read as a complete account of training variability. All 300 Magentic-NL scheduled pairs are customer-focal, so its customer-utility difference concerns this customer-focal panel. The sampled-futures model is retrained for each of the three seeds and paired with the corresponding \method{} run.

For judge validation, one author manually annotated all 100 interactions in the stratified sample for each of the four judged environments (400 interactions in total), with method identities hidden. We report Spearman correlation between ratings from the LLM judge and this author for SOTOPIA goal, AgentSense goal, ProMediate mediator intelligence and SOTOPIA-ToM InfoMgmt. Correlations range from 0.68 to 0.76, indicating partial agreement on these adapted metrics. Because a single author supplied the human ratings, inter-annotator agreement is not applicable. This is an internal validation rather than an independent assessment, and the correlations neither establish probability calibration nor eliminate judge bias. Deterministic Magentic-NL payoffs require no judge comparison.

\begin{table*}[tp]
\centering
\small
\setlength{\tabcolsep}{4pt}
\begin{tabularx}{\textwidth}{@{}X ccccccc@{}}
\toprule
Environment & Seeds & Pairs & $\Delta\uparrow$ [95\% CI] & \multicolumn{2}{c}{Failure (\%) $\downarrow$} & Judge $\rho\uparrow$ & Human $n$ \\
 & & & & \method{} & Samples & & \\
\midrule
Magentic-NL & 3 & 300 & \shortstack{$14.8$\\$[4.2,\,25.6]$} & 0.0 & 0.0 & n/a & n/a \\
SOTOPIA & 3 & 300 & \shortstack{$1.11$\\$[0.90,\,1.32]$} & 0.0 & 0.0 & 0.76 & 100 \\
AgentSense & 3 & 300 & \shortstack{$0.08$\\$[0.02,\,0.14]$} & 0.0 & 0.0 & 0.71 & 100 \\
ProMediate & 3 & 300 & \shortstack{$0.0043$\\$[-0.0011,\,0.0097]$} & 0.0 & 0.0 & 0.68 & 100 \\
SOTOPIA-ToM & 3 & 300 & \shortstack{$0.026$\\$[0.008,\,0.044]$} & 0.0 & 0.0 & 0.73 & 100 \\
\bottomrule
\end{tabularx}
\caption{Reliability audit for \method{} versus 30 sampled futures on a separate evaluation schedule. The audit uses 300 scheduled pairs across three training seeds per environment. All 300 pairs completed in each environment; both methods have zero episode failures. Differences and intervals describe all 300 completed pairs. Human $n$ counts interactions, all annotated by one author. Judge agreement is inapplicable to deterministic Magentic-NL payoffs. Primary measures, sampling and failure rules are specified in Appendix~\ref{app:analyses}.}
\label{tab:reliability}
\vspace{-1em}
\end{table*}

\begin{table*}[tp]
\centering
\small
\setlength{\tabcolsep}{3pt}
\begin{tabularx}{\textwidth}{@{}X ccccc@{}}
\toprule
Predictor & RMSE $\downarrow$ & Rank $\uparrow$ & Raw valid (\%) $\uparrow$ & Tokens / cand. $\downarrow$ & Latency (ms) $\downarrow$ \\
\midrule
Single future & \withsd{0.178}{0.008} & \withsd{0.49}{0.024} & \withsd{\underline{98.8}}{0.6} & \withsd{\underline{438}}{12} & \withsd{\underline{610}}{46} \\
Direct expectation & \withsd{0.134}{0.010} & \withsd{0.64}{0.029} & \cellcolor{resultgreen}\withsd{\textbf{99.6}}{0.6} & \cellcolor{resultgreen}\withsd{\textbf{112}}{3} & \cellcolor{resultgreen}\withsd{\textbf{185}}{13} \\
Sampled futures (30) & \withsd{0.151}{0.008} & \withsd{\underline{0.69}}{0.026} & \withsd{92.1}{0.6} & \withsd{13140}{328} & \withsd{6200}{474} \\
\method{} (SFT) & \withsd{\underline{0.133}}{0.008} & \withsd{\underline{0.69}}{0.027} & \withsd{96.8}{0.5} & \withsd{760}{17} & \withsd{1160}{96} \\
\method{} (SFT + GRPO) & \cellcolor{resultgreen}\withsd{\textbf{0.116}}{0.011} & \cellcolor{resultgreen}\withsd{\textbf{0.71}}{0.030} & \withsd{98.1}{0.5} & \withsd{812}{25} & \withsd{1280}{91} \\
\bottomrule
\end{tabularx}
\caption{Final-horizon SOTOPIA diagnostics on shared saved states, reported as mean $\pm$ SD across five seeds after averaging three backbones: RMSE of normalized expected outcome values ($\downarrow$), candidate-ranking agreement (Kendall's $\tau_b$, $\uparrow$), valid raw requests ($\uparrow$), generated tokens and latency per candidate. Accuracy uses the shared retained cases; costs include retries. Highlighting follows Table~\ref{tab:main-seen}.}
\label{tab:forecast-cost}
\vspace{-1em}
\end{table*}

\subsection{Mechanism Analyses}\label{app:mechanism-analysis}

\paragraph{Variation across seeds.}
The main tables and prediction diagnostics report mean $\pm$ SD. Within each seed, we average the three evaluation backbones using the same weights as the reported mean, and then compute the sample SD across the five seed-level results. This spread describes variation across the runs and is neither a standard error nor a confidence interval for a method difference. In particular, the three backbone evaluations of a trained checkpoint are not three independent training seeds. The separate reliability audit retains its three-seed protocol and scenario-cluster confidence intervals.

\paragraph{Paired comparisons.}
Table~\ref{tab:paired-uncertainty} reports differences for eight comparisons on the five-seed main evaluation grid. We compute paired method differences using the same aggregation weights as the main tables and average over the three evaluation backbones. To obtain 95\% confidence intervals, we use 10,000 bootstrap resamples over both scenarios and the five training/evaluation seeds, preserving pairing between methods. The intervals therefore account for variation across scenarios and seeds, conditional on the three evaluated backbones. The number of independent training seeds remains five, so increasing the resample count improves the numerical stability of the bootstrap without adding independent runs. We interpret each interval separately rather than as a simultaneous guarantee across all comparisons.

\paragraph{World modeling and post-training.}
The paired comparisons support gains from distributional world modeling in several environments, with the strongest evidence depending on the baseline. Under SFT alone, \method{} exceeds direct expectation on four primary measures with intervals above zero, while its AgentSense difference is $-0.030$ with an interval of $[-0.060,0.005]$. Full \method{} likewise exceeds direct expectation in four environments, but its AgentSense difference of $0.020$ remains inconclusive. Against sampled futures, full \method{} improves on all five primary measures with intervals above zero, including ProMediate ($0.0059$, $[0.0014,0.0132]$). GRPO improves over SFT alone in Magentic-NL, SOTOPIA, AgentSense and SOTOPIA-ToM; the ProMediate difference is only $0.0015$, with an interval of $[-0.0027,0.0058]$. Together, these comparisons support the contribution of both the trained future-state representation and post-training, while leaving some environment-specific advantages unresolved. The SFT-only comparison holds the training stage fixed, although prediction format and target aggregation still differ.

\paragraph{Information used for action selection.}
The representation comparisons show where the additional prediction information helps the actor. Full guidance exceeds expected values only in all four environments outside Magentic-NL, with intervals above zero. Keeping probabilities visible and adding response text improves SOTOPIA, AgentSense, ProMediate and SOTOPIA-ToM, with intervals above zero in all four. In turn, retaining response probabilities and residual-mass annotations without text improves over expected values only in AgentSense and ProMediate; the corresponding SOTOPIA and SOTOPIA-ToM intervals include zero. This latter comparison changes response probabilities and residual-mass annotations together, while expected outcome values and combined utility stay fixed. Full guidance also exceeds automatic utility selection in SOTOPIA, ProMediate and SOTOPIA-ToM, while the Magentic-NL and AgentSense intervals include zero. The zero Magentic-NL differences agree with the aggregate results under its payoff-only filter, but do not establish identical paired outcomes. Overall, richer future representations help in several settings, while the benefit of each component depends on the environment.

\begin{table*}[t]
\centering\footnotesize
\setlength{\tabcolsep}{3pt}
\begin{tabularx}{\textwidth}{@{}X ccccc@{}}
\toprule
 & Magentic-NL & SOTOPIA & AgentSense & ProMediate & SOTOPIA-ToM \\
Comparison & Cust.\ util. $\uparrow$ & Goal $\uparrow$ & Goal $\uparrow$ & Cons.\ chg. $\uparrow$ & InfoMgmt $\uparrow$ \\
\midrule
SFT $-$ direct expectation & \shortstack{$52.60$\\{\scriptsize $[43.10, 61.90]$}} & \shortstack{$0.240$\\{\scriptsize $[0.050, 0.430]$}} & \shortstack{$-0.030$\\{\scriptsize $[-0.060, 0.005]$}} & \shortstack{$0.0173$\\{\scriptsize $[0.0101, 0.0245]$}} & \shortstack{$0.0333$\\{\scriptsize $[0.0220, 0.0446]$}} \\[3pt]
Full $-$ direct expectation & \shortstack{$64.14$\\{\scriptsize $[55.20, 73.40]$}} & \shortstack{$0.620$\\{\scriptsize $[0.430, 0.820]$}} & \shortstack{$0.020$\\{\scriptsize $[-0.006, 0.046]$}} & \shortstack{$0.0188$\\{\scriptsize $[0.0114, 0.0262]$}} & \shortstack{$0.0668$\\{\scriptsize $[0.0554, 0.0781]$}} \\[3pt]
Full $-$ sampled futures & \shortstack{$17.26$\\{\scriptsize $[8.10, 26.40]$}} & \shortstack{$1.110$\\{\scriptsize $[0.900, 1.320]$}} & \shortstack{$0.130$\\{\scriptsize $[0.099, 0.161]$}} & \shortstack{$0.0059$\\{\scriptsize $[0.0014, 0.0132]$}} & \shortstack{$0.0348$\\{\scriptsize $[0.0239, 0.0458]$}} \\[3pt]
Full $-$ SFT & \shortstack{$11.54$\\{\scriptsize $[3.80, 19.20]$}} & \shortstack{$0.380$\\{\scriptsize $[0.210, 0.560]$}} & \shortstack{$0.050$\\{\scriptsize $[0.019, 0.081]$}} & \shortstack{$0.0015$\\{\scriptsize $[-0.0027, 0.0058]$}} & \shortstack{$0.0335$\\{\scriptsize $[0.0231, 0.0439]$}} \\[3pt]
Full $-$ without response text & \shortstack{$0.00$\\{\scriptsize $[-1.20, 1.10]$}} & \shortstack{$0.190$\\{\scriptsize $[0.006, 0.374]$}} & \shortstack{$0.090$\\{\scriptsize $[0.055, 0.125]$}} & \shortstack{$0.0054$\\{\scriptsize $[0.0009, 0.0115]$}} & \shortstack{$0.0160$\\{\scriptsize $[0.0052, 0.0269]$}} \\[3pt]
Full $-$ expected values only & \shortstack{$0.00$\\{\scriptsize $[-1.40, 1.30]$}} & \shortstack{$0.230$\\{\scriptsize $[0.060, 0.400]$}} & \shortstack{$0.120$\\{\scriptsize $[0.087, 0.153]$}} & \shortstack{$0.0117$\\{\scriptsize $[0.0051, 0.0183]$}} & \shortstack{$0.0202$\\{\scriptsize $[0.0088, 0.0316]$}} \\[3pt]
Without text $-$ expected values only & \shortstack{$0.00$\\{\scriptsize $[-1.10, 1.10]$}} & \shortstack{$0.040$\\{\scriptsize $[-0.120, 0.200]$}} & \shortstack{$0.030$\\{\scriptsize $[0.004, 0.056]$}} & \shortstack{$0.0063$\\{\scriptsize $[0.0009, 0.0117]$}} & \shortstack{$0.0042$\\{\scriptsize $[-0.0060, 0.0144]$}} \\[3pt]
Full $-$ automatic utility selection & \shortstack{$0.00$\\{\scriptsize $[-1.10, 1.10]$}} & \shortstack{$0.250$\\{\scriptsize $[0.080, 0.420]$}} & \shortstack{$0.030$\\{\scriptsize $[-0.002, 0.062]$}} & \shortstack{$0.0061$\\{\scriptsize $[0.0005, 0.0125]$}} & \shortstack{$0.0117$\\{\scriptsize $[0.0008, 0.0226]$}} \\[3pt]
\bottomrule
\end{tabularx}
\caption{Paired differences on the main evaluation grid. Each cell shows the difference between reported means (first method minus second) above its 95\% confidence interval from 10,000 bootstrap resamples over scenarios and five training/evaluation seeds. Full denotes \method{} (SFT + GRPO); SFT denotes \method{} (SFT only). Positive differences favor the first method. Intervals account for scenario and seed variation, conditional on the three evaluation backbones.}
\label{tab:paired-uncertainty}
\vspace{-1em}
\end{table*}

\paragraph{Prediction diagnostics.}
Table~\ref{tab:mechanism-diagnostics} compares predictions on shared held-out states and candidate actions in three training environments. RMSE measures normalized error in expected outcome values, while Kendall's $\tau_b$ compares candidate rankings under a common objective. Regret is the reference-estimated utility of the best candidate minus that of the candidate with highest predicted utility, before eligibility filtering. It therefore evaluates numerical action ranking rather than the observed outcomes after actor selection. Fifty reference continuations per candidate action supply the comparison values. Table~\ref{tab:forecast-input-counts} reports the input inventory before exclusions. All methods use the same retained 94\% of diagnostic cases, including repaired predictions, under the handling rules in Appendix~\ref{app:diagnostic-handling}. SFT-only comparisons hold the training stage fixed, although individual-response targets and cluster means still differ in aggregation and label noise.

\begin{table*}[tp]
\centering\small
\setlength{\tabcolsep}{3pt}
\begin{tabularx}{\textwidth}{@{}l X ccc@{}}
\toprule
Environment & Predictor & RMSE $\downarrow$ & $\tau_b\uparrow$ & Regret $\downarrow$ \\
\midrule
Magentic-NL & Single future & \withsd{0.201}{0.010} & \withsd{0.43}{0.038} & \withsd{0.118}{0.008} \\
 & Direct expectation & \withsd{0.171}{0.009} & \withsd{0.58}{0.027} & \withsd{0.086}{0.006} \\
 & Sampled futures (30) & \withsd{0.145}{0.009} & \withsd{0.66}{0.029} & \withsd{0.071}{0.006} \\
 & \method{} (SFT) & \withsd{\underline{0.139}}{0.010} & \withsd{\underline{0.68}}{0.032} & \withsd{\underline{0.065}}{0.007} \\
 & \method{} (SFT + GRPO) & \cellcolor{resultgreen}\withsd{\textbf{0.127}}{0.007} & \cellcolor{resultgreen}\withsd{\textbf{0.73}}{0.030} & \cellcolor{resultgreen}\withsd{\textbf{0.054}}{0.009} \\
\midrule
SOTOPIA & Single future & \withsd{0.178}{0.008} & \withsd{0.49}{0.024} & \withsd{0.094}{0.005} \\
 & Direct expectation & \withsd{0.134}{0.010} & \withsd{0.64}{0.029} & \withsd{0.069}{0.007} \\
 & Sampled futures (30) & \withsd{0.151}{0.008} & \withsd{\underline{0.69}}{0.026} & \withsd{0.063}{0.007} \\
 & \method{} (SFT) & \withsd{\underline{0.133}}{0.008} & \withsd{\underline{0.69}}{0.027} & \withsd{\underline{0.062}}{0.006} \\
 & \method{} (SFT + GRPO) & \cellcolor{resultgreen}\withsd{\textbf{0.116}}{0.011} & \cellcolor{resultgreen}\withsd{\textbf{0.71}}{0.030} & \cellcolor{resultgreen}\withsd{\textbf{0.057}}{0.007} \\
\midrule
AgentSense & Single future & \withsd{0.164}{0.013} & \withsd{0.56}{0.031} & \withsd{0.087}{0.007} \\
 & Direct expectation & \withsd{0.158}{0.014} & \withsd{0.62}{0.028} & \withsd{0.068}{0.007} \\
 & Sampled futures (30) & \withsd{0.169}{0.011} & \withsd{0.55}{0.030} & \withsd{0.091}{0.008} \\
 & \method{} (SFT) & \withsd{\underline{0.142}}{0.012} & \withsd{\underline{0.65}}{0.029} & \withsd{\underline{0.066}}{0.007} \\
 & \method{} (SFT + GRPO) & \cellcolor{resultgreen}\withsd{\textbf{0.135}}{0.011} & \cellcolor{resultgreen}\withsd{\textbf{0.66}}{0.045} & \cellcolor{resultgreen}\withsd{\textbf{0.064}}{0.009} \\
\bottomrule
\end{tabularx}
\caption{Final-horizon prediction diagnostics, mean $\pm$ SD across five seeds after averaging three evaluation backbones. RMSE and regret use normalized scales; regret concerns the highest predicted-utility candidate before filtering. All methods use the same retained cases (94\% of the diagnostic pool). Higher $\tau_b$ and lower RMSE and regret are better. Highlighting follows Table~\ref{tab:main-seen}.}
\label{tab:mechanism-diagnostics}
\vspace{-1em}
\end{table*}

\paragraph{Future representations.}
The expected-values-only representation uses the same final \method{} checkpoint and world modeling procedure as the full representation. It keeps visible context, candidate actions, immediate and final expected outcome values, combined utility, eligibility and fallback rules fixed at a given decision point. It removes response descriptions, their probabilities and residual-mass annotations, but it does not recompute utilities from expected values. The without-response-text variant instead retains the probability information. Table~\ref{tab:mechanisms} reports online outcomes, so histories can diverge after different choices. These comparisons vary what the actor reads rather than how the WM is trained.

\paragraph{Prediction diagnostic findings.}
Among SFT-only predictors in Table~\ref{tab:mechanism-diagnostics}, \method{} has higher rank agreement and lower regret than direct expectation in all three environments, and its mean RMSE is lower in each: 0.139 vs.\ 0.171 in Magentic-NL, 0.133 vs.\ 0.134 in SOTOPIA and 0.142 vs.\ 0.158 in AgentSense. The SOTOPIA error difference is small even though ranking agreement differs (0.69 vs.\ 0.64), and the AgentSense regret margin is also small (0.066 vs.\ 0.068). Because direct expectation learns means of raw outcome values whereas \method{} learns response clusters with conditional means, this comparison also changes supervision aggregation. Regret assesses numerical candidate ranking, not the observed outcomes after actor selection. In AgentSense, sampled futures has the worst mean on all three diagnostics and falls below the actor alone on both outcome measures (Table~\ref{tab:main-seen}); this association is not causal.

\paragraph{Effects of the future representation.}
In Table~\ref{tab:mechanisms}, the without-response-text and expected-values-only variants keep the final checkpoint, expected outcome values, utilities and eligibility fixed, so they change only what the actor reads. Removing response text lowers the primary outcome metrics in four environments. Hiding probabilities and residual mass as well lowers these four metrics further. For example, AgentSense goal falls from 0.72 with full guidance to 0.63 without text and 0.60 with numerical guidance only. In contrast, Magentic-NL customer utility remains $272.82\pm7.91$ across both representation variants, full guidance and automatic utility selection. This agrees with its payoff-only dominance rule: only candidates with maximal predicted payoff remain eligible, and a unique maximum leaves no choice for the future representation to influence. Thus, these results provide no evidence of a benefit from the richer future representation in Magentic-NL. The paired intervals in Table~\ref{tab:paired-uncertainty} support a response-text benefit in four environments and a benefit from retaining response probabilities and residual-mass annotations without text in two. These comparisons do not estimate the aggregate effect of hiding probabilities while retaining text: that condition is examined only in the separate qualitative case in Figure~\ref{fig:decision-example}. Online histories can diverge after different choices, so the results concern the effects of the representations over complete interactions.

\paragraph{Nonlinear selection.}
Dominance filtering based on expected outcome values is a heuristic: an action with a better vector of expected outcome values need not have higher expected nonlinear utility. In particular, it can remove an action whose cluster-averaged geometric utility exceeds that of the retained actions. We do not quantify the frequency or utility cost of these conflicts. Such an audit would require the candidate predictions before and after filtering, with constraint eligibility held fixed.

\begin{figure}[t]
\centering
\includegraphics[width=\columnwidth]{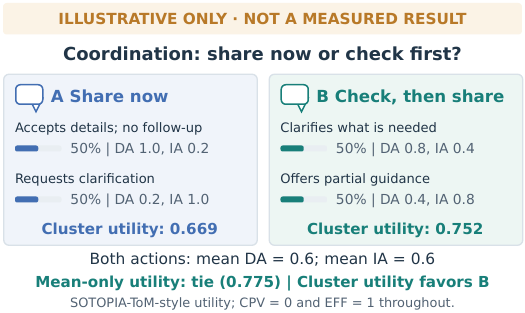}
\caption{\textbf{Hypothetical utility example.} Equal expected outcome values need not imply equal cluster-averaged nonlinear utility. The calculation illustrates aggregation, not an observed decision.}
\label{fig:aggregation-example}
\vspace{-1em}
\end{figure}

\paragraph{Illustrative decision example.}
Figure~\ref{fig:aggregation-example} uses the SOTOPIA-ToM-style geometric utility with CPV $=0$ and EFF $=1$. Action A has equally likely conditional mean outcome values $(\mathrm{DA},\mathrm{IA})=(1.0,0.2)$ and $(0.2,1.0)$; action B has $(0.8,0.4)$ and $(0.4,0.8)$. Both have marginal means $(0.6,0.6)$, so average-first utility is $(0.6\cdot0.6)^{1/4}=0.775$. Combining first gives $0.2^{1/4}=0.669$ for A and $0.32^{1/4}=0.752$ for B. Neither mean vector dominates the other. Thus, the illustration concerns aggregation order alone, not superior prediction accuracy or the causal value of response text. All actions, responses and values are hypothetical; the figure illustrates the calculation rather than an observed actor choice.

\paragraph{Qualitative comparison of future representations.}\label{app:illustrated-display}
Figure~\ref{fig:decision-example} examines a SOTOPIA scene in which Mia Davis wants quiet for her noise-sensitive dog, while Sophia James requests tree trimming before her daughter visits. The saved state and four candidate actions originate from turn 1 of a historical evaluation episode (seed 3). The figure abbreviates the actions as postponing pruning, trimming at night, arranging a certified arborist at a quieter time, or promising to handle the work. The scene is not tagged SOTOPIA-hard. The historical seed identifies the source state, not the separate run used for this comparison. The qualitative evaluation uses the best checkpoint from each training seed.

The separate qualitative run requests two response clusters per action, whereas the main evaluation requests four. For brevity, each panel shows prediction information only for its selected action; the actor evaluates all four candidates. The two predicted probabilities per action sum to one, and both are included whenever probabilities are shown. In this run, the actor using the full representation selected C and described it as a concrete plan addressing safety and noise. C also produced the best observed overall outcome, whereas D had the highest predicted utility. Without probabilities, the actor selected A, emphasizing quiet and the proposed shed repair. Without response text, it selected B, which had the highest expected goal value. With the expected-values-only representation, it selected D, which had the highest combined utility. The qualitative runs use the same selection prompt as the main evaluation (Appendix~\ref{app:selection}). The actor produces reasoning separately from its final candidate ID; Figure~\ref{fig:decision-example} records that reasoning, while action execution uses the final ID. The reasoning comes from the actor selecting an action, rather than the 4B WM predicting consequences. It explains the reported choice without establishing the internal cause of that choice. Run-specific actor and checkpoint identifiers and the separate-run seed are not tabulated here.

All four future representations retain the same numerical guidance and eligibility. Full guidance includes response text and probabilities; the other representations hide probabilities, response text, or both. Expected values are not recomputed when probability labels are hidden. The shown final goal/relationship pairs are $(6.4,3.7)$, $(7.2,3.0)$, $(7.0,3.2)$ and $(6.8,3.6)$ for A--D, with reported utilities 0.716, 0.717, 0.717 and 0.720. Other outcome dimensions and immediate values are omitted from the drawing. Each action trades off goal and relationship, so none dominates another on these two dimensions. All four remain eligible. The response probabilities are WM predictions, not empirical acceptance rates in this case.

The shown values, choices, replies and explanations were verified against the separate run. The local source record documents the historical state and actions, but it does not contain the outputs of that separate run. The reported outcome comparison identifies C as best according to observed overall outcome, whereas the numerical guidance in the figure consists of predicted final goal and relationship values and predicted utility. These quantities serve different purposes: the predictions guide the decision, and the observed overall outcome evaluates its result. Per-action realized overall values are not supplied here. Thus, the case documents a successful choice but does not establish an average advantage, how often masking changes decisions, or whether an explanation faithfully identifies the cause. The without-probabilities variant has no aggregate row in Table~\ref{tab:mechanisms}. This case therefore complements the aggregate representation comparisons without replacing them.

\section{Probability Perturbation and Measure Diversity}\label{app:sensitivity}

Tables~\ref{tab:probability-perturbation} and~\ref{tab:measure-diversity} report mean $\pm$ SD across five training/evaluation seeds, each averaging the three evaluation backbones. Figure~\ref{fig:sensitivity} plots these results relative to the unperturbed or full-inventory mean, respectively. Both means and SDs are divided by that fixed reference mean and multiplied by 100; the bands describe variation across seeds rather than confidence intervals for paired differences.

\paragraph{Probability perturbation.}
We mix the predicted probability vector $p$ with a random permutation $q$ of its entries, using $p_\alpha=(1-\alpha)p+\alpha q$. Response texts and conditional mean outcome values stay fixed, while expected values, utilities and eligibility are recomputed. Table~\ref{tab:probability-perturbation} reports achieved JS divergence from the original prediction, pooled across environments on common held-out inputs. These pooled values give the horizontal positions in Figure~\ref{fig:probability-sensitivity}; they do not imply identical divergence in each environment. This experiment measures sensitivity to probability assignments rather than calibration against an empirical reference distribution.

\begin{table*}[t]
\centering\footnotesize
\setlength{\tabcolsep}{3pt}
\begin{tabular*}{\textwidth}{@{\extracolsep{\fill}}ccccccc@{}}
\toprule
 & & Magentic-NL & SOTOPIA & AgentSense & ProMediate & SOTOPIA-ToM \\
$\alpha$ & JS & Cust.\ util. $\uparrow$ & Goal $\uparrow$ & Goal $\uparrow$ & Cons.\ chg. $\uparrow$ & InfoMgmt $\uparrow$ \\
\midrule
0.00 & 0.00 & \cellcolor{resultgreen}\withsd{\textbf{272.82}}{7.91} & \withsd{\underline{8.21}}{0.20} & \cellcolor{resultgreen}\withsd{\textbf{0.720}}{0.023} & \cellcolor{resultgreen}\withsd{\textbf{0.0628}}{0.0047} & \cellcolor{resultgreen}\withsd{\textbf{0.2508}}{0.0119} \\
0.25 & 0.04 & \withsd{\underline{265.73}}{8.24} & \cellcolor{resultgreen}\withsd{\textbf{8.23}}{0.22} & \withsd{\underline{0.700}}{0.025} & \withsd{\underline{0.0622}}{0.0049} & \withsd{\underline{0.2460}}{0.0124} \\
0.50 & 0.10 & \withsd{251.27}{8.86} & \withsd{8.02}{0.24} & \withsd{0.680}{0.028} & \withsd{0.0602}{0.0052} & \withsd{0.2347}{0.0131} \\
0.75 & 0.18 & \withsd{230.81}{9.42} & \withsd{7.73}{0.27} & \withsd{0.650}{0.030} & \withsd{0.0581}{0.0056} & \withsd{0.2202}{0.0143} \\
1.00 & 0.28 & \withsd{209.53}{10.17} & \withsd{7.36}{0.30} & \withsd{0.610}{0.033} & \withsd{0.0536}{0.0060} & \withsd{0.2021}{0.0155} \\
\bottomrule
\end{tabular*}
\caption{Probability perturbation: mean $\pm$ SD across five seeds after averaging three backbones. JS is achieved divergence from the original prediction, pooled across environments. The five primary outcome measures match Table~\ref{tab:mechanisms}; higher is better. Highlighting follows Table~\ref{tab:main-seen}.}
\label{tab:probability-perturbation}
\vspace{-1em}
\end{table*}

\begin{table*}[t]
\centering\footnotesize
\setlength{\tabcolsep}{3pt}
\renewcommand{\withsd}[2]{\shortstack{#1\\{\scriptsize\color{black}$\pm$#2}}}
\begin{tabular*}{\textwidth}{@{\extracolsep{\fill}}lcccccccc@{}}
\toprule
 & \multicolumn{3}{c}{ProMediate} & \multicolumn{5}{c}{SOTOPIA-ToM} \\
\cmidrule(lr){2-4}\cmidrule(lr){5-9}
Measures & \shortstack{Cons.\ chg.\\$\uparrow$} & \shortstack{Topic eff.\\$\uparrow$} & \shortstack{Med.\ intel.\\$\uparrow$} & \shortstack{InfoMgmt\\$\uparrow$} & DA $\uparrow$ & IA $\uparrow$ & CPV $\downarrow$ & EFF $\uparrow$ \\
\midrule
5 (20\%) & \withsd{0.0506}{0.0062} & \withsd{0.0030}{0.0006} & \withsd{3.5742}{0.244} & \withsd{0.1851}{0.0176} & \withsd{0.4512}{0.0311} & \withsd{0.1286}{0.0187} & \withsd{0.3041}{0.0356} & \withsd{0.8214}{0.0412} \\
10 (40\%) & \withsd{0.0546}{0.0058} & \withsd{0.0032}{0.0005} & \withsd{3.8147}{0.223} & \withsd{0.2079}{0.0158} & \withsd{0.4807}{0.0297} & \withsd{0.1423}{0.0172} & \withsd{0.2870}{0.0311} & \withsd{0.8538}{0.0376} \\
15 (60\%) & \withsd{0.0582}{0.0054} & \withsd{0.0035}{0.0005} & \withsd{4.0613}{0.205} & \withsd{0.2280}{0.0144} & \withsd{0.5085}{0.0279} & \withsd{0.1551}{0.0159} & \withsd{0.2674}{0.0254} & \withsd{0.8846}{0.0340} \\
20 (80\%) & \withsd{\underline{0.0610}}{0.0049} & \withsd{\underline{0.0037}}{0.0004} & \withsd{\underline{4.2418}}{0.193} & \withsd{\underline{0.2423}}{0.0128} & \withsd{\underline{0.5284}}{0.0256} & \withsd{\underline{0.1652}}{0.0139} & \withsd{\underline{0.2519}}{0.0210} & \withsd{\underline{0.9041}}{0.0319} \\
25 (100\%) & \cellcolor{resultgreen}\withsd{\textbf{0.0628}}{0.0047} & \cellcolor{resultgreen}\withsd{\textbf{0.0038}}{0.0004} & \cellcolor{resultgreen}\withsd{\textbf{4.3321}}{0.185} & \cellcolor{resultgreen}\withsd{\textbf{0.2508}}{0.0119} & \cellcolor{resultgreen}\withsd{\textbf{0.5398}}{0.0243} & \cellcolor{resultgreen}\withsd{\textbf{0.1695}}{0.0128} & \cellcolor{resultgreen}\withsd{\textbf{0.2450}}{0.0173} & \cellcolor{resultgreen}\withsd{\textbf{0.9138}}{0.0300} \\
\bottomrule
\end{tabular*}
\caption{Training-measure diversity: mean $\pm$ SD across five seeds after averaging three backbones. The inventory contains 25 outcome measures; training supervision is matched across conditions. Higher is better except CPV ($\downarrow$). Highlighting follows Table~\ref{tab:main-seen}.}
\label{tab:measure-diversity}
\vspace{-1em}
\end{table*}

\paragraph{Diversity of training outcome measures.}
The training inventory contains 25 outcome measures. We retain 5, 10, 15, 20 or 25 measures (20\%, 40\%, 60\%, 80\% or 100\%), restricting both SFT and GRPO to each selected inventory. Nested measure subsets and shared underlying rollouts keep the comparison tied to the same interaction data. We match the number of requested measures per training example and the training steps across conditions, including the full inventory. Every condition uses the full inference queries on the two unseen environments. Thus, this comparison varies the breadth of outcome supervision while keeping its volume fixed.

\end{document}